%% file: document_elsevier.tex
\documentclass[authoryear]{elsarticle}

\usepackage{hyperref}
\usepackage{amsmath}
\usepackage{amsfonts}
\usepackage{graphicx, xcolor}
\usepackage{booktabs}
\usepackage{array, makecell}
\usepackage{siunitx}
\usepackage{amssymb}
\usepackage{array, makecell}
\usepackage{graphicx, subfigure}
\usepackage[export]{adjustbox}
\usepackage{cleveref}
\usepackage{siunitx}
\usepackage{cleveref}
\usepackage{tabu}
\usepackage{caption}

\journal{arXiv}

\begin{document}

\begin{frontmatter}

\title{GANs for Medical Image Analysis}

\author[tud]{Salome Kazeminia\fnref{equal_contrib}}
\ead{salome.kazeminia@gris.tu-darmstadt.de}

\fntext[equal_contrib]{The authors contributed equally to this work.}

\author[tum]{Christoph Baur\fnref{equal_contrib}}
\ead{c.baur@tum.de}

\author[igd]{Arjan Kuijper}
\ead{arjan.kuijper@mavc.tu-darmstadt.de}

\author[bram]{Bram van Ginneken}
\ead{bram.vanginneken@radboudumc.nl}

\author[tum]{Nassir Navab}
\ead{navab@cs.tum.edu}

\author[tum]{Shadi Albarqouni}
\ead{shadi.albarqouni@tum.de}

\author[tud]{Anirban Mukhopadhyay}
\ead{anirban.mukhopadhyay@gris.tu-darmstadt.de}

\address[tud]{Department of Computer Science, TU Darmstadt, Germany}
\address[tum]{Computer Aided Medical Procedures (CAMP), TU Munich, Germany}
\address[igd]{Fraunhofer IGD, Darmstadt, Germany}
\address[bram]{Radboud University Medical Center, Nijmegen, The Netherlands}




\begin{abstract}
	\input{abstract}
\end{abstract}


\end{frontmatter}

%\linenumbers

	\input{sections/introduction}

	\input{sections/mediaproblems}
	\input{sections/overviews}
	%	
	\input{sections/applications}

	\input{sections/discussion}

	\section*{References}
	
	\bibliographystyle{plainnat}
	\bibliography{literature}

\end{document}

%% file: abstract.tex
Generative Adversarial Networks (GANs) and their extensions have carved open many exciting ways to tackle well known and challenging medical image analysis problems such as medical image de-noising, reconstruction, segmentation, data simulation, detection or classification. Furthermore, their ability to synthesize images at unprecedented levels of realism also gives hope that the chronic scarcity of labeled data in the medical field can be resolved with the help of these generative models. In this review paper, a broad overview of recent literature on GANs for medical applications is given, the shortcomings and opportunities of the proposed methods are thoroughly discussed and potential future work is elaborated. We review the most relevant papers published until the submission date. For quick access, important details such as the underlying method, datasets and performance are tabulated. An interactive visualization which categorizes all papers to keep the review alive, is available at \url{http://livingreview.in.tum.de/GANs_for_Medical_Applications/}. 

%% file: sections/introduction.tex
\section{Introduction}
\label{sec:introduction}

From the early days of Medical Image Analysis, Machine Learning (ML) and Artificial Intelligence (AI) driven systems have been a key component for complex decision making - a brief history of which can be found in \citep{litjens2017survey}. 
Across generations of development, the focus was mostly on decision making at different granularity levels. 
These techniques range from low-level pixel processing over feature engineering combined with supervised classifier learning to the recent wave of feature learning using Convolutional Neural Networks (CNNs).  

Until recently, the driving focus of the Medical Image Analysis (MIA) community has been on the supervised learning of decision boundaries, while generative tasks have been on the back seat. This changed dramatically with the advent of Generative Adversarial Networks (GANs) ~\citep{goodfellow2014generative}, which lead to a new age of generative modeling and distribution learning. With their abilities to mimic data distributions and to synthesize images at yet unprecedented levels of realism, GANs have carved open new ways to bridge the gap between \textit{supervised learning} and \textit{image generation}.

The mentioned improvements are essentially due to the following properties: 1) GANs maximize the probability density over the data generating distribution by exploiting density ratio estimation \citep{Isola2017ImagetoImageTW} in an indirect fashion of supervision; 2) GANs can discover the high dimensional latent distribution of data, which has lead to significant performance gains in the extraction of visual features. 

This review summarizes GAN-based architectures proposed for medical image processing applications published until the end of 2018.  

\begin{figure}[t]
	\centering
	\includegraphics[width=1\textwidth]{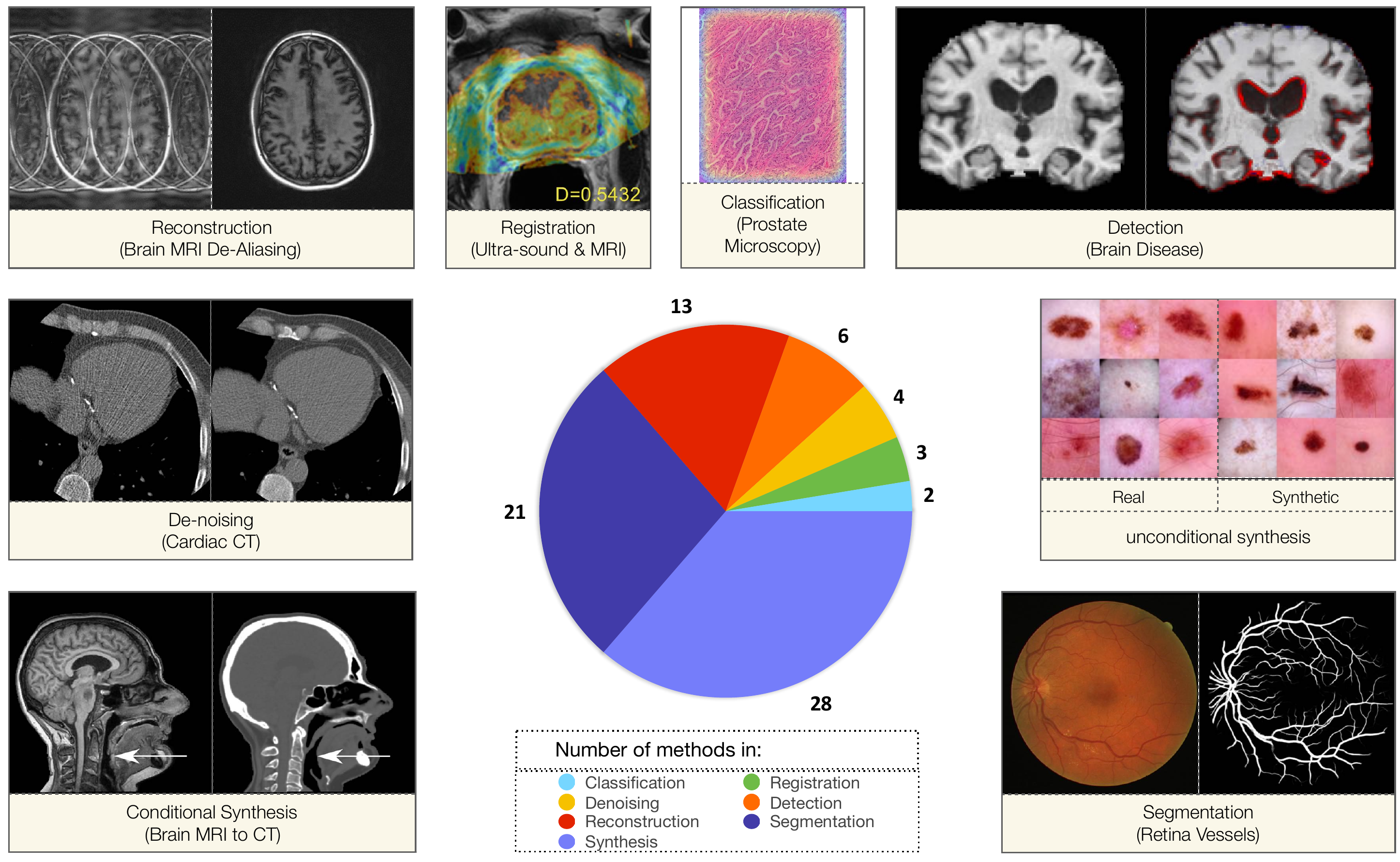}
	\caption{The pie chart of distribution of papers and visual examples of GAN functionality among the different applications. Examples are taken from papers as the following: Conditional synthesis \cite{wolterink2017deep}, Denoising \cite{wolterink2017generative}, Reconstruction \cite{zhang2018multi}, Registration \cite{Yan2018AdversarialIR}, Classification \cite{Ren2018AdversarialDA}, Detection \cite{baumgartner2017visual}, Unconditional synthesis \cite{baur2018melanogans}, and Segmentation \cite{son2017retinal}.}
	\label{fig:pgan}
\end{figure}

We categorized the reviewed papers into 7 categories according to the following applications: synthesis, segmentation, reconstruction, detection, de-noising, registration, and classification. The categorical distribution of the reviewed papers can be seen in Figure \ref{fig:pgan}. Methods based on GANs were applied to a variety of different medical imaging modalities such as MRI (Magnetic Resonance Imaging), CT (Computed Tomography), OCT (Optical Coherence Tomography), chest X-Ray, Dermoscopy, Ultrasound, PET (Positron Emission Tomography) and Microscopy.

To find the papers we searched for the keywords “medical” and “GAN” (or “generative adversarial network”) along with the aforementioned applications in Google Scholar, Semantic Scholar, PubMed, and CiteSeer. Also, we checked references and citations of selected papers. Since GANs are rather new, and a significant number of articles are still in the publication process of different journals and conferences, we covered pre-prints published in arXiv as well.

For quick access, we summarized key information on the reviewed methods, such as metrics, datasets, image modality and the employed architectures in Tables. Also, we provide a live tree at \url{http://livingreview.in.tum.de/GANs_for_Medical_Applications/} for ease of categorization of papers. Further, we thoroughly discuss the advantages and shortcomings of the methods and specify clear directions for future work.

We thus ended up with 77 papers which we consider the most relevant ones covering a broad spectrum of applications and variety of GANs. The remainder of this paper is structured as follows:
 
In Section \ref{sec:overview}, we introduce the architecture of the GAN and its subclasses which are used in medical image applications. In Section \ref{sec:applications_in_medical_image_processing}, different contributions of GANs in medical image processing applications (de-noising, reconstruction, segmentation, registration, detection, classification, and synthesis) are described and Section \ref{sec:discussion} provides a conclusion about the investigated methods, challenges and open directions for the employment of GANs in medical image processing.

%% file: sections/mediaproblems.tex
\section{Opportunities for Medical Image Analysis}
\label{sec:mediaproblems}

% Addressing point 2) of Reviewer 1

Supervised Deep Learning is currently the state of the art in many Computer Vision and Medical Image Analysis tasks, but its success is heavily dependent on the large-scale availability of labeled training data. 
In the medical field, this is a fundamental problem as often there is a severe lack of labeled data. 
%data scarcity
This \emph{data scarcity} arises from the tedious, time-consuming and costly nature of medical image acquisition and labeling. 
%class imbalances
Furthermore, in the medical field, many datasets suffer from severe \emph{class imbalance} due to the rare nature of some pathologies. 
GANs can potentially alleviate these two limitations by generating realistic-looking images from an implicit distribution that follows the real data distribution.
The community has recognized this potential early and has been eagerly investigating GAN's suitability for tackling these problems. 
Applied research on GANs for such medical image synthesis can be broadly classified into two categories: i) Unconditional and ii) Conditional Image Synthesis. 
The GAN is originally an unconditional, unsupervised generative model which generates data from noise and leaves the user with limited influence on the generated output (a review of such methods can be found in Subsection \ref{sub:unsupervised_image_synthesis}). 
Conditional variants have also been proposed, which allow to seed the generation process with prior information such as class labels, image attributes or even images themselves (see Subsection \ref{sub:conditional_image_synthesis}).

While labeled data might be scarce in the medical field, unlabeled data is often readily available. 
To leverage both labeled and arbitrary amounts of unlabeled data, numerous so-called \emph{Semi-Supervised Deep Learning} methods have been proposed throughout the last years. 
Recent advances in this field make use of adversarial networks, the core-concept behind GANs, to harness both labeled and unlabeled data in the training of classifiers and segmentation models.
%domain shift
A related issue hampering Deep Learning is the effect of so-called \emph{domain shift}, i.e. a shift between the training and testing data distribution. 
Such a shift frequently occurs in the medical field and often leads to models that generalize poorly beyond the training data, with potentially unpredictable behavior. 
Prominent examples are i) MR data, where images from different device vendors exhibit very different characteristics, or ii) histopathological images, which vary greatly due to different staining protocols and equipments. 
GAN-based Domain Adaptation methods have recently shown great potential to close this gap, some of which are reviewed in Subsection \ref{sub:Segmentation} and \ref{sub:classification}.

This potential comes primarily from the introduction of an adversarial network into the overall optimization process. According to \cite{odena2016conditional}, such a network can be understood as a learned, rich similarity metric: Traditional similarity metrics such as the $\ell_1$ or $\ell_2$-distance are known to produce blurry results and lack the incorporation of spatial context, whereas such a learned similarity metric allows to optimize for concepts in images beyond the pixel level. This appealing property has recently been exploited in medical image denoising (see Subsection \ref{sub:denoising}), superresolution, image-to-image translation (see Subsection  \ref{sub:conditional_image_synthesis}) and even medical image segmentation (see Subsection \ref{sub:Segmentation}).

%time efficiency
The newly achieved levels of realism promoted by adversarial networks also open up new opportunities for tackling problems with high time-complexity, such as medical image reconstruction or registration. 
These problems are mathematically well modeled, but usually require a costly iterative optimization. 
GANs have been successfully used to learn a mapping from raw signals to reconstructed images (see Subsection \ref{sub:reconstruction}) or to register medical images (see Subsection \ref{sub:Registration}) in a single forward pass while ensuring coherence and high levels of realism.

%% file: sections/overviews.tex
\section{Basic GAN models employed in medical applications}
\label{sec:overview}

In this section, we introduce the general concept behind GANs, their conditional variants as well as a variety of prominent extensions. 
These extensions comprise the DCGAN, Markovian GAN, conditional GAN, CycleGAN, Auxiliary Classifier GAN, Wasserstein-GAN as well as Least Squares GAN which are used as the inspiring models in medical applications.

In the context of this work, there are three \textit{“Adversarial”} concepts, which should be understood properly by their different meanings. 
\textit{“Adversarial attack”} means to make imperceptible changes to an image such that a classifier misclassifies it. 
Usually, the modified image, called “adversarial image” or \textit{“adversarial examples”} \citep{goodfellow2014explaining}, is not distinguishable from the original image visually \citep{Szegedy2013IntriguingPO}. 
\textit{“Adversarial training”} proposed by  \cite{Szegedy2013IntriguingPO} is an idea that increases the robustness of neural networks against adversarial attacks by training the model with both normal and adversarial examples.
Due to the state of existing neural networks at that time, implementing adversarial training was not a practical solution. The effectiveness of this idea became apparent when Goodfellow et. al employed it in GANs \citep{goodfellow2014generative}. 

\subsection{GAN}
\label{Overview:GAN}
The GAN framework~\citep{goodfellow2014generative} consists of a training dataset $\mathbf{X}$, whose underlying distribution we denote $p_{real}$, and a pair of competing networks: a generator (G) with parameters $\theta_{G}$, and a discriminator (D) with weights $\theta_{D}$ (see Figure \ref{fig:GAN}).
G aims to find a mapping $\hat{x} = G(z;\theta_{G})$ that maps latent random variables $z \sim p_z(z)$---drawn from a prior distribution $p_z$---to generated data $\hat{x} \in \mathbf{\hat{X}}$, which is supposed to follow the distribution $p_{\theta}(\hat{x}|z)$.
The primary goal is to optimize this mapping such that the distribution of generated data $\mathbf{\hat{X}}$ resembles the distribution of the training data $\mathbf{X}$, i.e. $p_{\theta}(\hat{x}|z) \sim p_{real}$. In other words, G is supposed to generate fake data which must not be distinguishable from real data. 
This is achieved with the help of the discriminator network D, whose task is to classify between fake and real samples. Essentially, D is a binary classifier which yields $D(x) = 1$ for real samples and $D(\hat{x}) = 0$ for fake data. 
Both networks are adversaries as G attempts to gradually synthesize more and more realistic samples which D would misclassify as real, while D constantly learns to differentiate between real and synthesized samples. 
Mathematically speaking, D and G play a two-player minimax game with the following value function $V(G, D)$:  
\begin{equation}
	\label{eq:minmax}
	%\underset{G}{min}\underset{D}{max}V(D,G) = \mathbb{E}_{x\sim p_{data}(x)}[log D(x)]+\mathbb{E}_{z\sim p_{z}(z)}[log(1-D(G(z)))]
	\min_{G} \max_{D} V(D,G) = \mathbb{E}_{x \sim p_{data}(x)}[log(D(x))] + \mathbb{E}_{z \sim p_z(z)}[1-log(D(G(z)))]
\end{equation}
To optimize for Equation \ref{eq:minmax}, D is trained to maximize the probability of correct label assignment for fake and real data, while G is trained to trick D into thinking a generated sample is real by minimizing $log(1-D(G(z)))$. 
In practice, these networks are implemented as Multi-Layer Perceptrons (MLPs) and usually trained with minibatch stochastic gradient descend in an alternating fashion. 
Once trained, it is sufficient to sample a random $z$ and feed it through the generator in order to synthesize data.
This \textit{“adversarial training”} framework exhibits a few interesting properties: 1) the generator G is updated only through gradients back-propagated from the discriminator and 2) no explicit correspondences between $z$, $x$ and $\hat{x}$ are required, such that input is not explicitly memorized by G. 3) Furthermore, as proven by the authors, optimization for Equation \ref{eq:minmax} minimizes the Jensen-Shannon (JS) divergence between the distributions of real and synthetic data.

Although theoretically well grounded, the vanilla GAN has proven to be quite hard to train. General convergence is heavily dependent on hyper-parameter tuning to avoid vanishing or exploding gradients, and they are prone to \emph{mode collapse}. This term describes a phenomenon where GANs map all $z$ to very similar synthetic samples covering only a single mode of the data distribution. During optimization, these modes might also change (so-called \emph{mode hopping}).
A plethora of extensions and subclasses have been proposed to cope with these problems, a selection which is introduced in the following subsections.

\subsection{DCGAN:}	
\label{overview:DCGAN}
To address the instability of the basic GAN architecture and increase the resolution of synthesized images, \cite{Radford2015UnsupervisedRL} propose the Deep Convolutional GAN (DCGAN).
In this model, both the generator and discriminator follow a deep convolutional network architecture, exploiting the efficacy of spatial kernels and hierarchical feature learning. Concepts such as Batch-Normalization and Leaky-ReLU have been included to improve training stability, but issues such as mode collapse were not entirely resolved.

\subsection{cGAN:}
\label{overview:GAN:cGAN}
\begin{figure}
	\centering	
%	\medskip	
	\begin{subfigure}
		\centering\includegraphics[width=.8\textwidth, right]{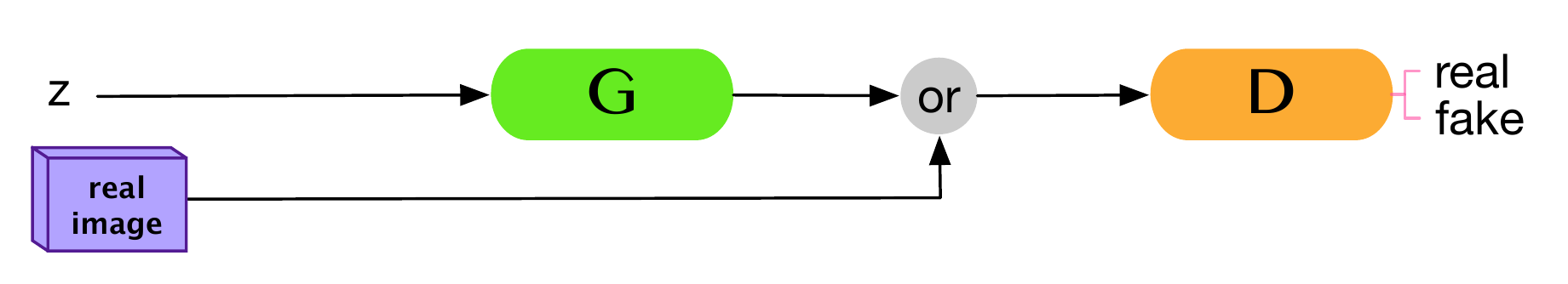}			
		\caption{GAN\label{fig:GAN}}		
	\end{subfigure}
%	\bigskip\hrulefill\bigskip
	\medskip	 
	\begin{subfigure}
		\centering\includegraphics[width=.8\textwidth, right]{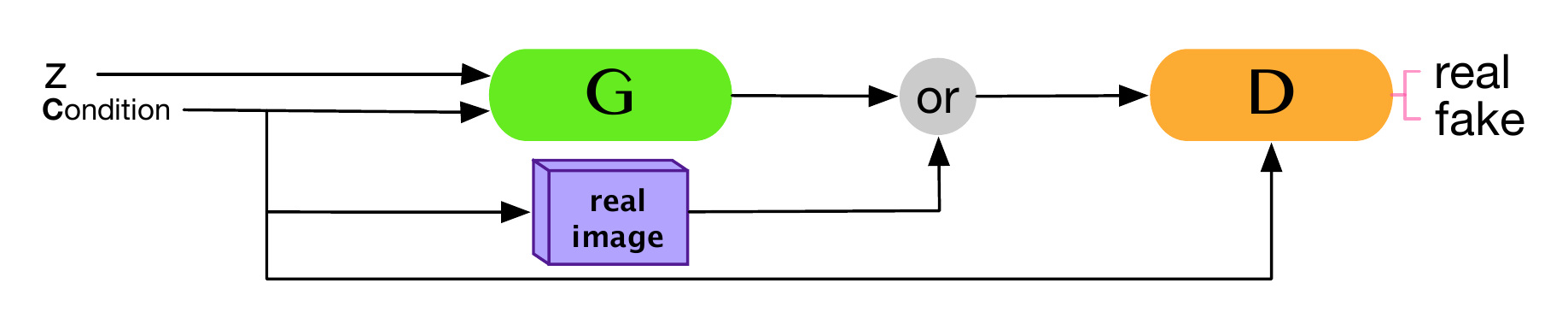}		
		\caption{cGAN\label{fig:cGAN}}
	\begin{subfigure}
		\centering\includegraphics[width=.8\textwidth, right]{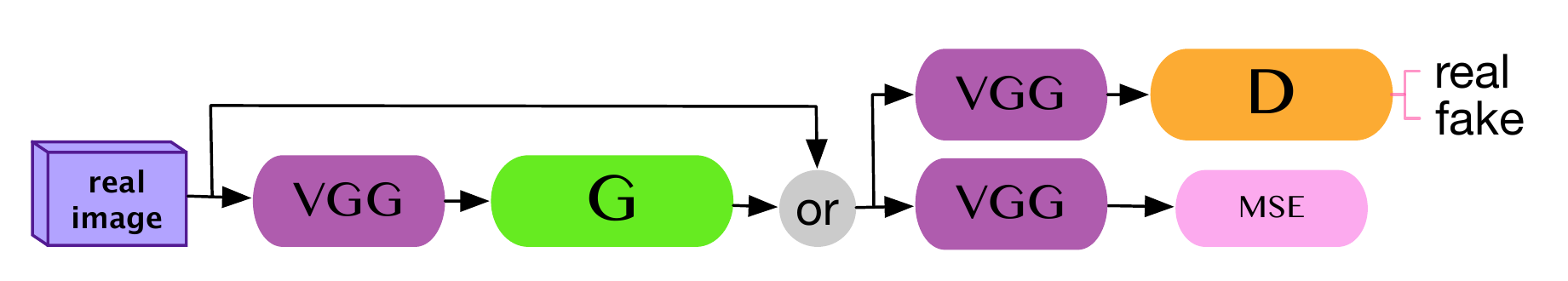}		
		\caption{MGAN\label{fig:MGAN}}
	\end{subfigure}
%	\bigskip\hrulefill\bigskip
	\medskip 
	\end{subfigure}
%	\bigskip\hrulefill\bigskip
    \medskip      
	\begin{subfigure}
		\centering\includegraphics[width=1\textwidth, right]{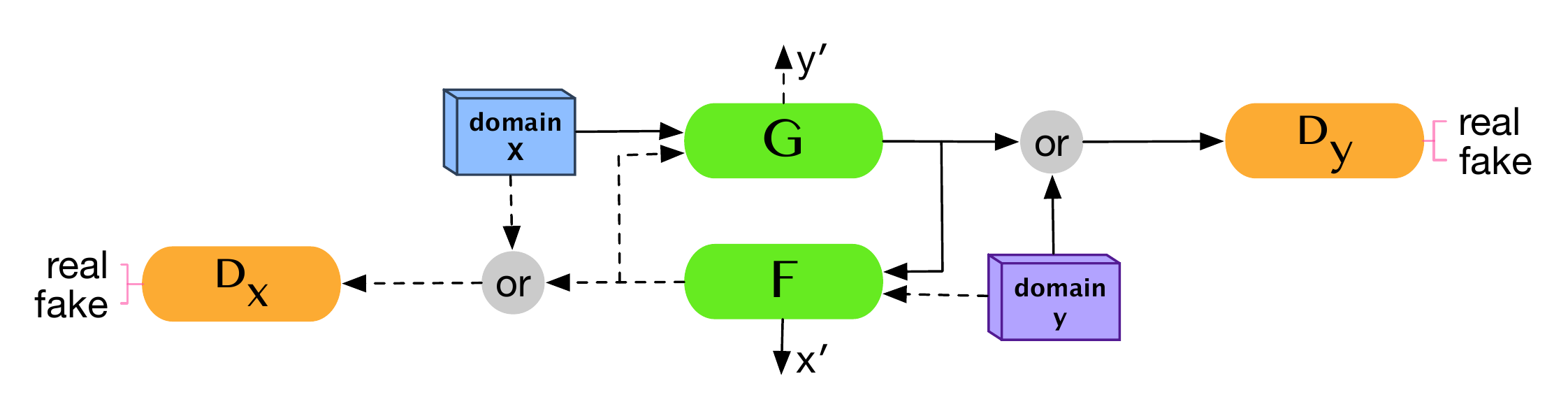}		
		\caption{cycleGAN\label{fig:cycleGAN}}
	\end{subfigure}	
%	\bigskip\hrulefill\bigskip
	\medskip
	\begin{subfigure}
		\centering\includegraphics[width=.8\textwidth, right]{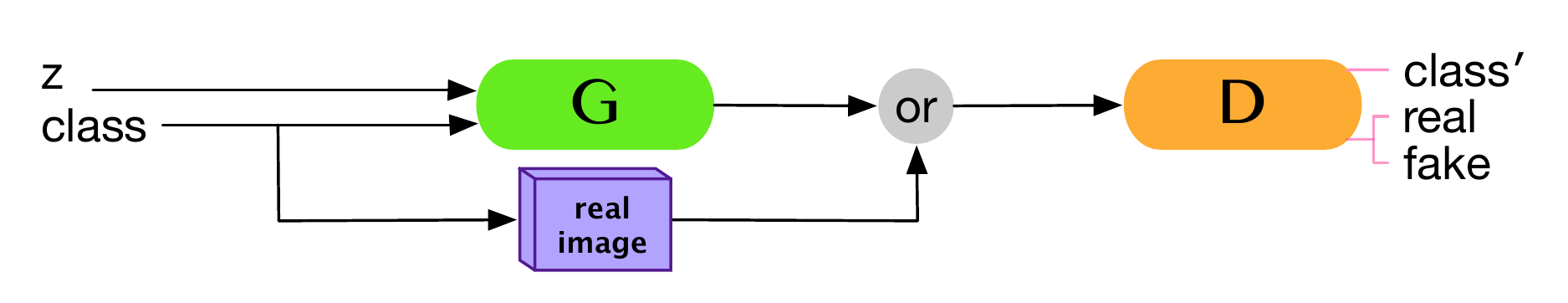}
		\caption{AC-GAN\label{fig:ACGAN}}
	\end{subfigure}			
\end{figure}

Since in the original GAN no control on the actual image generation is given, \cite{mirza2014conditional} proposed the conditional GAN (cGAN) to incorporate additional information like class labels in the image generation process (Figure \ref{fig:cGAN}).
In the cGAN, the generator is presented with random noise $z$ jointly with some prior information $c$. Additionally, the prior knowledge $c$ is fed into the discriminator together with the corresponding real or fake data. 
Mathematically speaking, the cGAN framework is given as follows:
\begin{equation}
	\min_{G} \max_{D} V(D,G) = \mathbb{E}_{x \sim p_{data}(x)}[log(D(x|c))] + \mathbb{E}_{z \sim p_z(z)}[1-log(D(G(z|c)))]
\end{equation}	
The authors showed that conditioning the GAN not only improves the generation of detailed features in the target image but also helps training stability.

Another conditional GAN framework is the Markovian GAN (MGAN) \citep{li2016precomputed}, proposed for fast and high-quality style transfer. 
The MGAN, as depicted in Figure \ref{fig:MGAN}, heavily utilizes a pre-trained VGG19 network with fixed weights to extract high-level features for both style transfer and preserving the image content. Both the discriminator and generator network are prepended with a VGG19 network to extract feature maps. An additional perceptual loss on those extracted feature maps forces the generator to generate images with realistic VGG19 feature activations--as would have been obtained on real data.

\cite{Isola2017ImagetoImageTW} proposed Pix2Pix, a very successful variant of the cGAN for high-resolution image-to-image translation. The Pix2Pix generator follows the U-Net \citep{ronneberger2015u} architecture, while the discriminator--similar to MGAN--follows a fully convolutional architecture for differentiating between the real and fake high resolution data.
The authors showed that the skip connections within the U-Net generator are very beneficial for global coherence of the synthesized images. In contrast to the original GAN framework, Pix2Pix requires pairs of corresponding input and desired output images, though. This allows the use of the $\ell_1$ loss between the generators' output and actual ground-truth image to stabilize the training.

\subsection{cycleGAN:}
\label{overview:GAN:cycleGAN}
For image transformation between two domains, the model should be able to extract characteristic features of both domains and to discover the underlying relationship between those.
\cite{Zhu2017UnpairedIT} proposed the CycleGAN to provide these principles.
In essence, the framework combines two GANs to find a mapping from domain X to domain Y and vice versa. These consist of a generator $G:X \rightarrow Y$, trained by discriminator $D_{Y}$, and generator $F:Y \rightarrow X$, trained by discriminator $D_{X}$ (Figure \ref{fig:cycleGAN}). 
The two GANs are chained together and a cyclic loss function forces them to reduce the space between their possible mapping functions. More precisely, this cyclic loss function minimizes the discrepancy between the original image and the reconstruction obtained from the chained generators.
The final loss function of cycleGAN is defined as:
\begin{equation}
	L(G,F,D_{X},D_{Y}) = L_{GAN}(G,D_{Y},X,Y) + L_{GAN}(F,D_{X},Y,X)+ \lambda L_{cyc}(G,F)
\end{equation}
with
\begin{equation}
	L_{cyc}(G,F) = \mathbb{E}_{x\sim P_{data}(x)}[\left \| F(G(x))-x \right \|_{1}]+\mathbb{E}_{y\sim P_{data}(y)}[\left \| G(F(y))-y \right \|_{1}]
\end{equation}

\subsection{AC-GAN}
\label{overview:acgan}	
The Auxiliary Classifier GAN (AC-GAN) proposed by \cite{odena2016conditional} describes a different approach towards constructing a GAN conditioned on classes. 
Authors report that instead of providing both the generator and the discriminator networks with prior information (as shown in the cGAN), the discriminator can be additionally tasked with respectively classifying its input. 
More precisely, the discriminator architecture is modified such that after a few layers it splits into a standard discriminator network as well as an auxiliary classifier network (see Figure \ref{fig:ACGAN}), which aims at classifying samples into different categories.
According to the authors, this allows to use (partially) pre-trained discriminators and appears to stabilize training.

\subsection{WGAN:}
\label{overview:GAN:WGAN}
In the discussed frameworks, the data distributions of generated and real images are matched by means of the Jensen-Shannon (JS) divergence. This divergence measure potentially makes gradients vanishing and the saddle-point of optimization unreachable, which are the underlying reasons behind GAN failures. \cite{Arjovsky2017WassersteinG} proposed the Wasserstein-GAN (WGAN) which uses the Earth Mover (ME) or Wasserstein-1 distance as a more optimal divergence measure to avoid vanishing gradients. 
This model has proven to be much more robust, while easy to implement. In practice, the downside of the WGAN is its slow optimization.

\subsection{LSGAN:}
\label{overview:GAN:LSGAN}
\cite{Mao2017LeastSG} also tried to tackle the training instability of GANs with their so-called Least Squares GAN (LSGAN). 
Similarly to WGAN, the loss function is modified to avoid vanishing gradients:
\begin{equation}
	%	\underset{D}{min} V_{LSGAN}(D)=\frac{1}{2}\mathbb{E}_{x\sim p_{data}(x)}[(D(x)-b)^{2}]+\frac{1}{2}\mathbb{E}_{z\sim p_{z}(z)}[(D(G(z))-a)^{2}]
	\min_{G} \max_{D} V(D,G) = \mathbb{E}_{x \sim p_{data}(x)}[(D(x)-b)^2] + \mathbb{E}_{z \sim p_z(z)}[(D(G(z))-a)^2]
\end{equation}
This loss-function operates directly on the logits of the discriminator network, where $a=0$ and $b=1$ are the controlling parameters for fake and real data, respectively.
This way, the fake images which are discriminated as real, even if they are far away from the dense distribution of real data, will be penalized more during optimization. 
On the other hand, the gradient faces the least value only if the distribution of fake data perfectly matches the distribution of real data.

		%\begin{equation}
		%\underset{G}{min} V_{LSGAN}(G)=\frac{1}{2}\mathbb{E}_{z\sim p_{z}(z)}[(D(G(z))-c)^{2}]
		%\end{equation}

%% file: sections/applications.tex
\section{Applications in Medical Image Processing}
\label{sec:applications_in_medical_image_processing}

In this section, we summarize GAN-based methods which are proposed to solve medical imaging problems in 7 application categories: synthesis, segmentation, reconstruction, detection, de-noising, registration and classification. 
In every subsection, a table summarizes the most important details in reviewed methods: the basic architectures and loss functions proposed, image modality and dataset properties, the evaluated performance of the proposed method, and if the paper underwent peer-review (PR) or provided any source code.
Since various metrics are used for evaluation of different methods, we provided Table \ref{table:metric} to explain metrics briefly. 

\begin{table}[!ht]
	\tiny
	\centering\renewcommand\cellalign{lc}
	\caption{Metric explanation. (seg: segmentation, rec: reconstruction, det: detection, den: de-noising, reg: registration, and clas: classification) (TP: true positive, TN: true negative, FP: false positive, and FN: false negative)}
	\label{table:metric}
	\begin{adjustbox}{width=\textwidth}{%
			\begin{tabu} to \textwidth {@{}X[0.6l] X[1.1l] X[0.5l] X[0.7l] X[1.1l] X[0.5l]@{}}
				\toprule
				Abbreviation & Metric & App &	Abbreviation & Metric & Application \\ 
				\hline
				
				ROC & Receiver Operating Characteristic &seg &AUC &Area under the ROC Curve & \makecell{seg, \\det}
				\\ \hline
				
				IoU &\makecell{Intersection over Union: \\$\frac{area of overlap}{area of union}$} &seg &Accuracy & $\frac{TP+TN}{TP+FP+FN+TN}$ &\makecell{seg, \\clas}
				\\ \hline
				
				Precision &$\frac{TP}{TP+FP}$ &\makecell{seg, \\clas} &Specifity &$\frac{TN}{TN+FP}$ & det
				\\ \hline
				
				DSC &\makecell{Dice similarity coefficient\\$2\frac{TP}{2TP+FP+FN}$} &\makecell{seg, \\det, \\reg} &Sensitivity &$\frac{TP}{TP+FN}$ &\makecell{seg, \\det}
				\\ \hline
				
				Recall &Sensitivity &\makecell{seg, \\det, \\clas} &F-score &$2\frac{Precision \times Recall}{Precision + Recall}$ &seg
				\\ \hline
				
				ASD &age symmetric surface distance &seg &MSD &mean surface distance &seg
				\\ \hline
				
				HSD &mean hausdorff surface distance &seg &OH &object hausdorff &seg
				\\ \hline
				
				correlation of EF & correlation of ejection fraction &seg &MAE of Infarct size &mean absolute error on the surface of infarction & seg
				\\ \hline
				
				NCC &normalized cross correlation &det &correct detection &(subjective) &det
				\\ \hline
				
				MSE &mean squared error &rec &NMSE &normalized MSE& rec
				\\ \hline
				
				RMSE &root MSE &rec &NRMESE &normalized root MSE &rec
				\\ \hline
				
				TRE &target registration error &rec &SIS &semantic interpretability score &seg
				\\ \hline
				
				SNR &signal to noise ratio &rec &PSNR &peak SNR &\makecell{rec, \\den}
				\\ \hline
				
				MOS &mean opinion score (subjective) &rec &GCF &global contrast factor &rec
				\\ \hline
				
				SSIM &structural similarity &\makecell{rec, \\den} &Agatston score &clinical intensity weighted measure of calcified area &den
				\\ \hline
				
				Noise suppression &scores in [1, 5](subjective) &den &Artifact reduction &scores in [1, 5](subjective) &den
				\\ \hline
				
				Overal Quality &scores in [1, 5](subjective) &den & P2PE &point-to-point errors & den
				\\ 
				\bottomrule
			\end{tabu}%
		}
	\end{adjustbox}
\end{table}

\input{sections/applications/synthesis}

\input{sections/applications/segmentation}

\input{sections/applications/reconstruction}

\input{sections/applications/detection}

\input{sections/applications/denoising}

\input{sections/applications/Registration}

\input{sections/applications/classification}

\begin{figure}
	\centering	
	%	\medskip
	\begin{subfigure}
		\centering\includegraphics[width=0.7\linewidth]{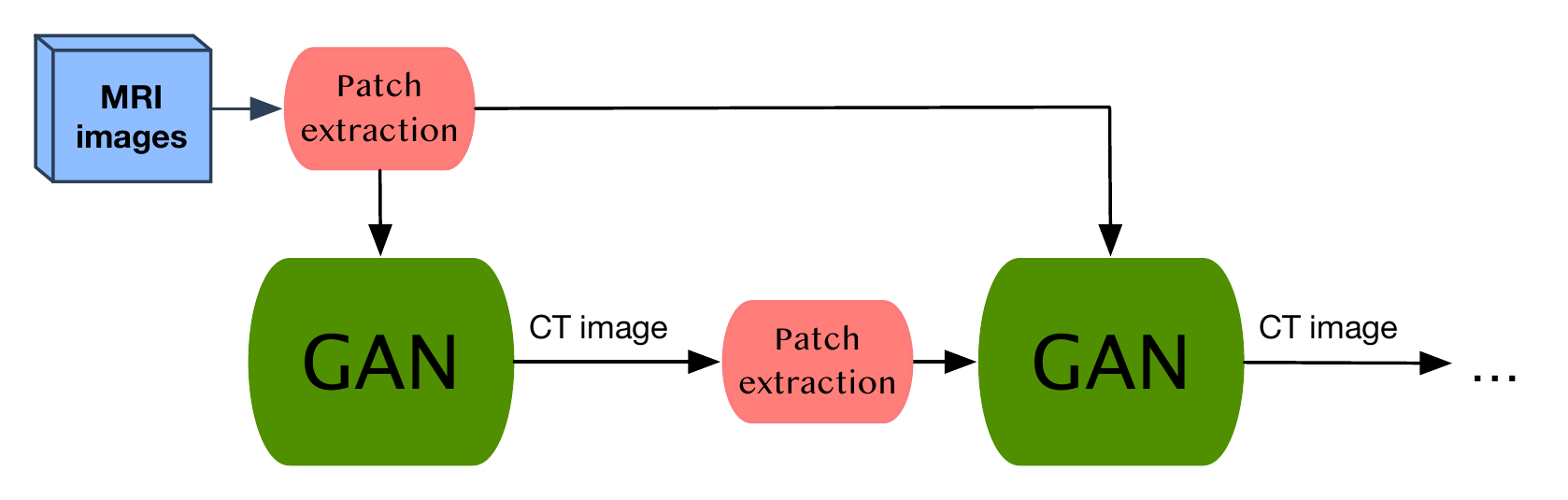}
		\caption{Synthesis:Proposed architecture by \cite{quan2017compressed}}
		\label{fig:syn-nie2017}
	\end{subfigure}	
	%	\bigskip\hrulefill\bigskip
	\medskip
	\begin{subfigure}
		\centering\includegraphics[width=0.7\linewidth]{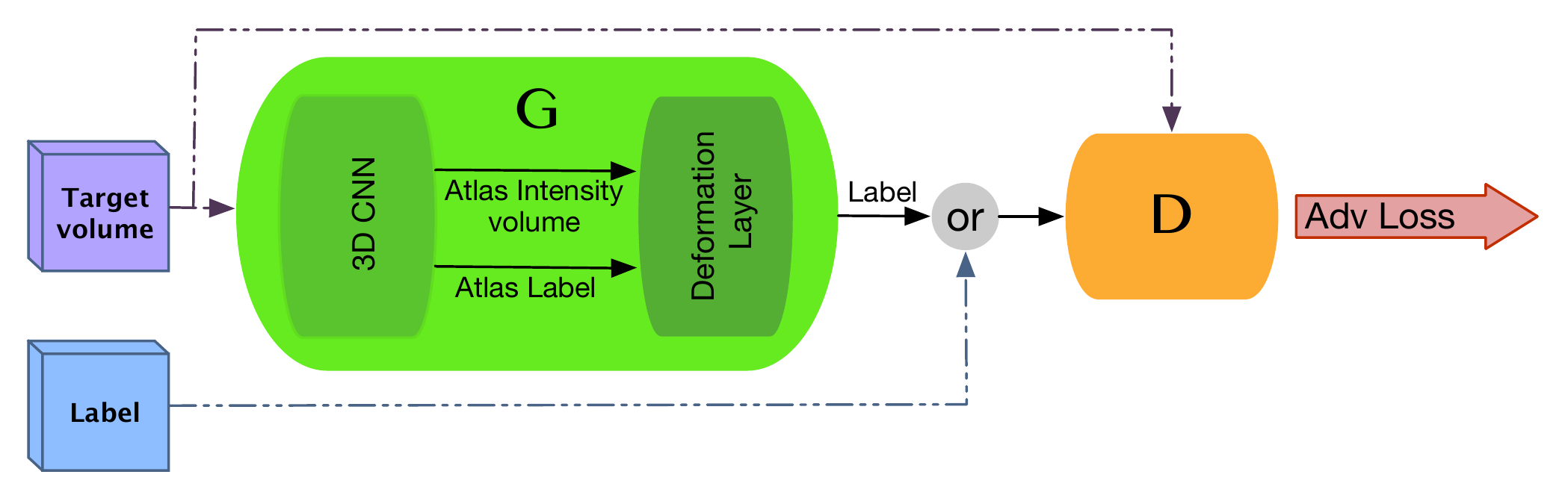}
		\caption{Segmentation: Proposed architecture by \cite{Dong2018VoxelAtlasGAN3L}}
		\label{fig:seg1}
	\end{subfigure}
	%	\bigskip\hrulefill\bigskip
	\medskip
	\begin{subfigure}
		\centering\includegraphics[width=0.6\linewidth]{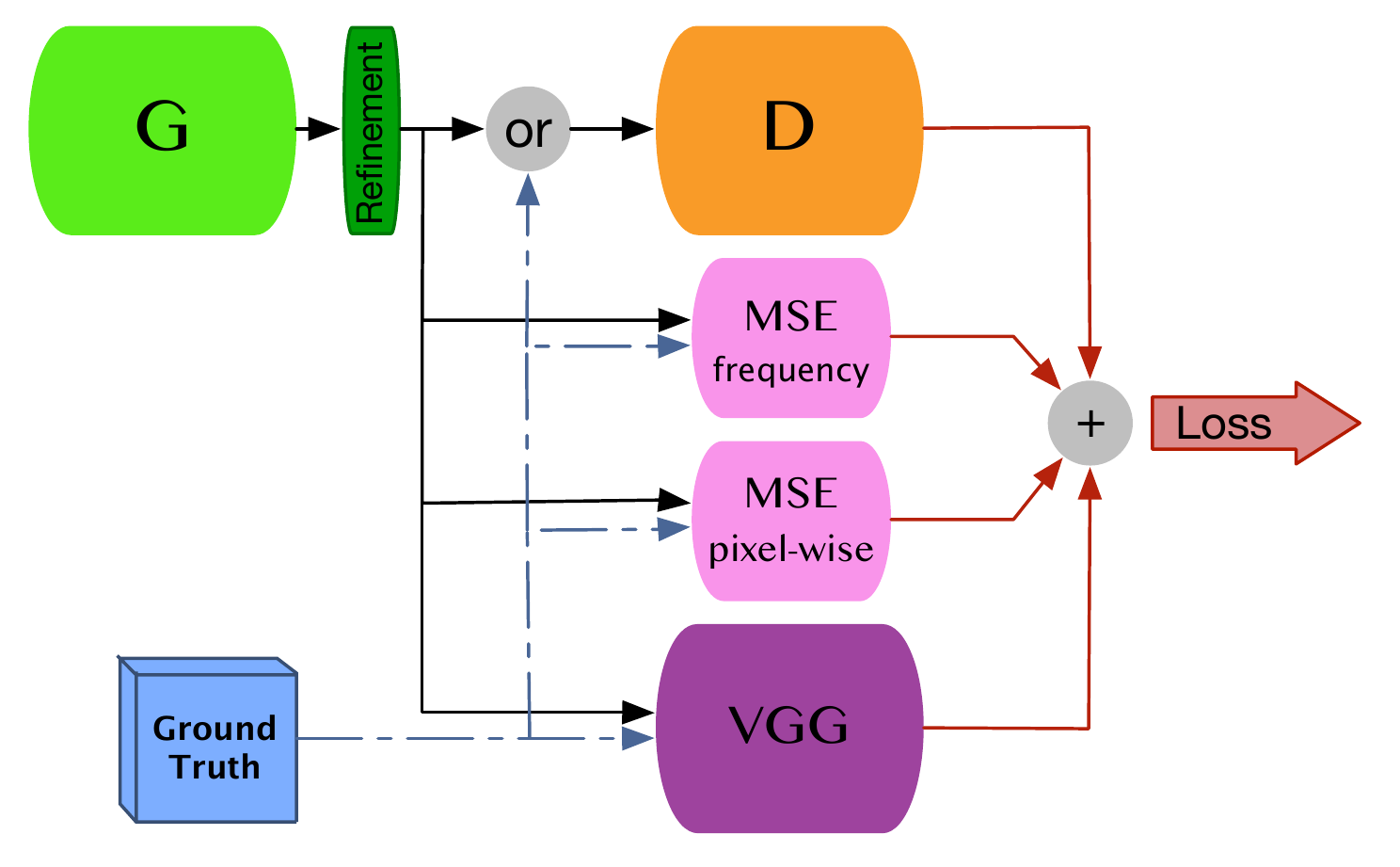}
		\caption{Reconstruction: DAGAN architecture \citep{yang2017dagan}}
		\label{fig:recons1}		
	\end{subfigure}
	%	\bigskip\hrulefill\bigskip
	\medskip	 
	\begin{subfigure}
		\centering\includegraphics[width=0.8\linewidth]{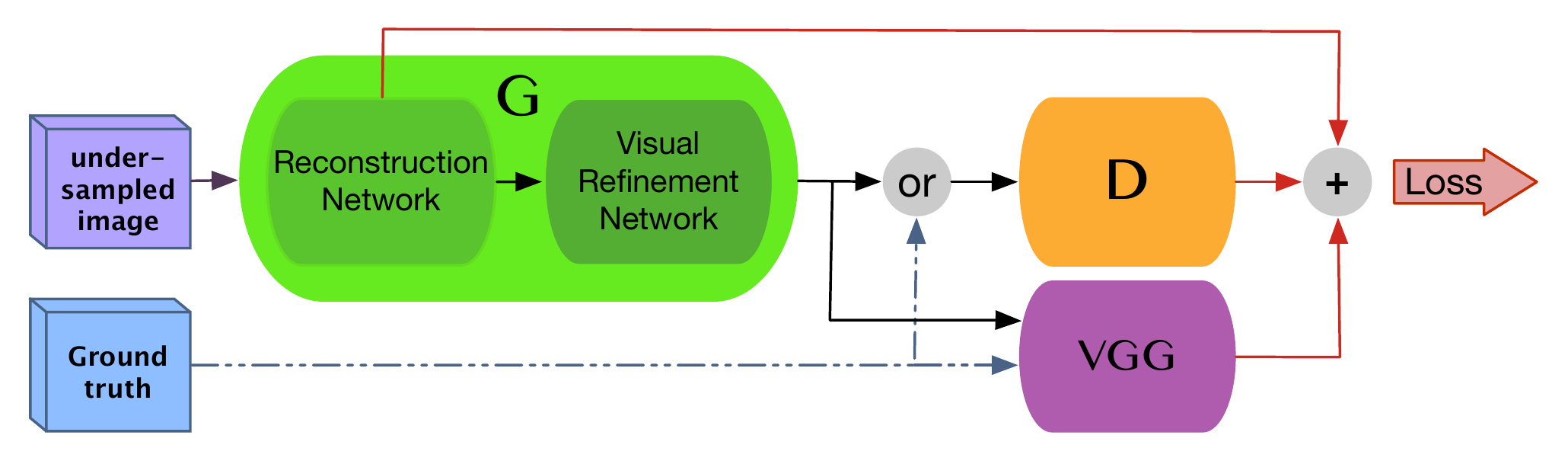}
		\caption{ReconstructionProposed architecture by \cite{seitzer2018adversarial}}
		\label{fig:recons3}
	\end{subfigure}
\end{figure}
\begin{figure}[!h]\ContinuedFloat
	\centering
	\begin{subfigure}
		\centering\includegraphics[width=0.7\linewidth]{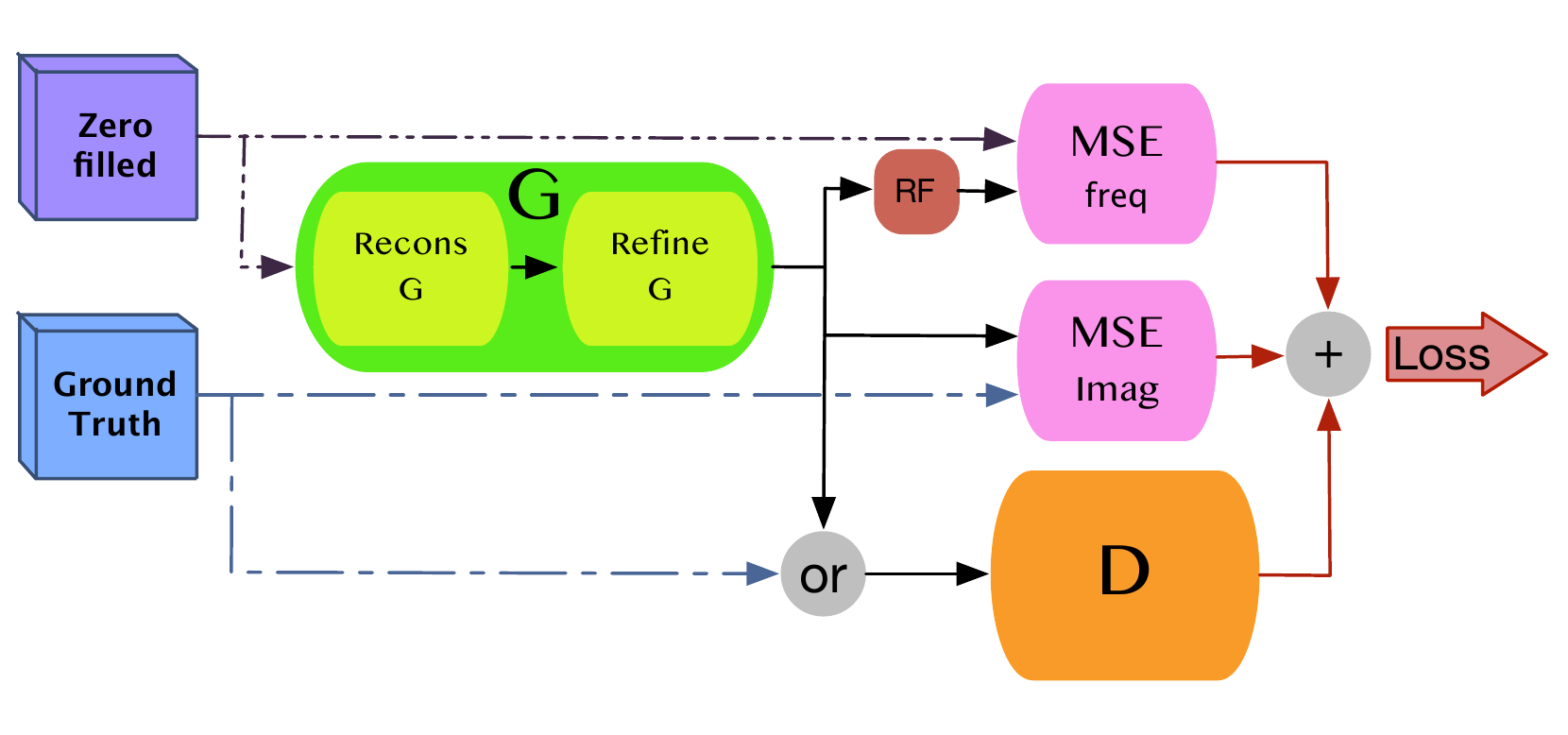}
		\caption{Reconstruction: RefineGAN architecture \citep{quan2017compressed}}
		\label{fig:recons2}
	\end{subfigure}
	%	\bigskip\hrulefill\bigskip
	\medskip
	\begin{subfigure}
		\centering\includegraphics[width=0.7\linewidth]{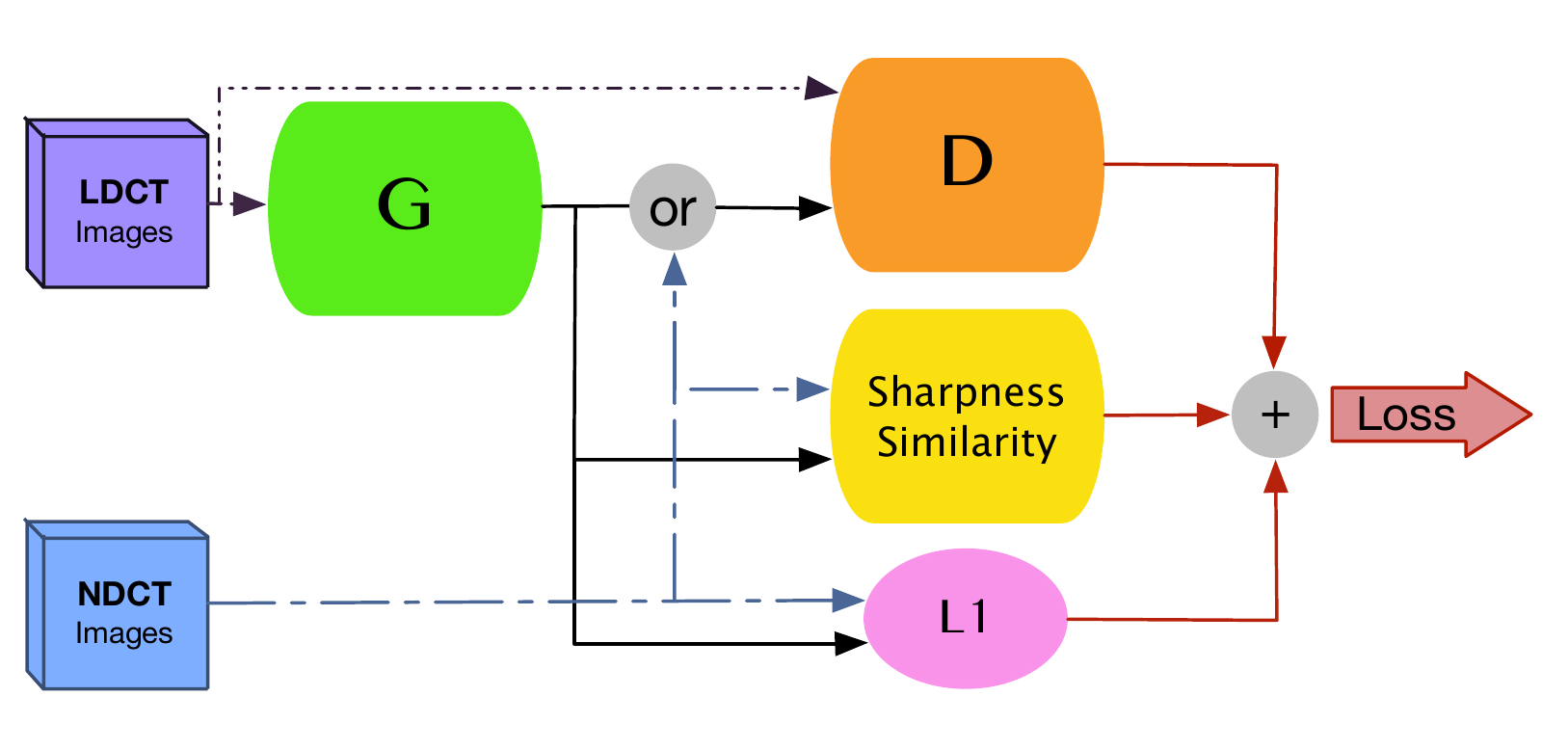}
		\caption{Denoising: SAGAN architecture \citep{Yi2018SharpnessAwareLC}}
		\label{fig:denoise4}
	\end{subfigure}	
	%	\bigskip\hrulefill\bigskip
	\medskip
	\begin{subfigure}
		\centering\includegraphics[width=0.7\linewidth]{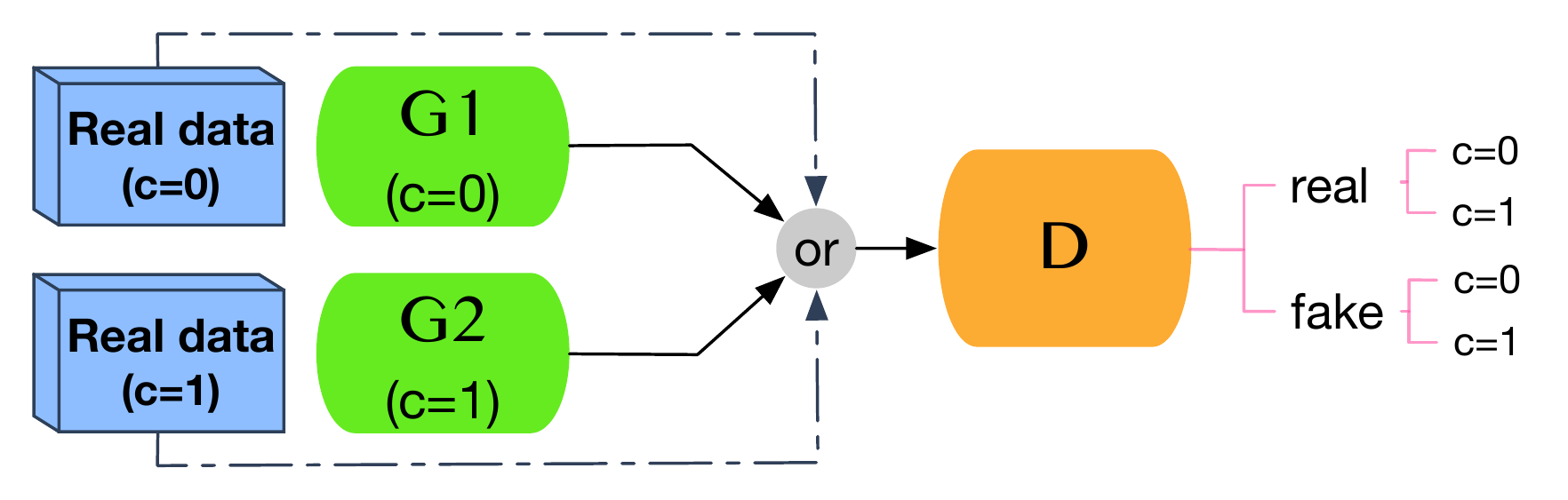}
		\caption{Classification: SCGAN architecture \citep{zhang2017semi}}
		\label{fig:class1}
	\end{subfigure} 	
\end{figure}

%% file: sections/applications/synthesis.tex
\subsection{Synthesis}
\label{sub:synthesis}

Originally, GANs have been proposed as an entirely unsupervised generative framework, with the ability to map from random noise to synthetic, realistically looking images.
With the conditional GAN, the framework has also been successfully turned into a supervised generative framework by conditioning on prior knowledge, rather than noise alone. 
For clarity, we refer to the original GAN framework as the \emph{unconditional} or \emph{unsupervised} GAN, in contrast to the \emph{conditional} GAN.
The generative property of both frameworks has been exploited in various ways for synthesizing certain types of medical images. 
In the following, a broad overview on works from both categories will be given. 
In the particular case for conditional approaches, we further classify the contributions based on the image modality.

\subsubsection{Unconditional Image Synthesis:}
\label{sub:unsupervised_image_synthesis}		
A great variety of works has recently appeared in the field of unsupervised medical image generation using GANs, which allows to tackle problems such as data scarcity and class imbalance \citep{frid2018synthetic}, facilitates data simulation \citep{chuquicusma2017fool} and aids to gain deeper insights into the nature of data distributions and their latent structure.

Initial results have shown that the DCGAN can be used to synthesize realistically looking, small patches of prostate lesions \citep{kitchen2017deep}, retinal images \citep{schlegl2017unsupervised} or lung cancer nodules \citep{chuquicusma2017fool}. 
The synthesized lung cancer nodules could even hardly be distinguished from real patches in a visual turing test involving two radiologists. \cite{frid2018synthetic} make also use of the DCGAN, but for the synthesis of focal CT liver lesion patches from different classes. 
For each class, i.e. cysts, metastases and hemangiomas, they train a separate generative model. As the training dataset is originally quite small, they used heavily augmented data to train the GANs. 
In a set of experiments for liver lesion classification, the authors demonstrate that synthetic samples---in addition to data augmentation---can considerably improve a CNN classifier.

\cite{bermudez2018learning} have shown that the DCGAN is also able to learn to mimic the distribution of entire MR data at considerably high resolution, even from a surprisingly small amount of samples. After training for 1500 epochs, the authors obtained visually compelling results which human observers could not reliably distinguish from real MR midline slices. 

\cite{baur2018melanogans} compare the DCGAN, LAPGAN and some LAPGAN-modifications for skin lesion synthesis at high resolution. 
Due to the high variance within the training data, the small number of samples turned out not to be sufficient to train a reliable DCGAN. However the hierarchical LAPGAN and its variants showed promising synthesis results. 
The synthetic samples have also been successfully used in the training of a skin lesion classifier. 
\cite{baur2018generating} employed the recently proposed concept of progressive GAN growing \citep{karras2017progressive} to synthesize high resolution images of skin lesions and showed stunning, highly realistic synthetic images which expert dermatologists could not reliably tell apart from real samples.

\subsubsection{Conditional Image Synthesis:}
\label{sub:conditional_image_synthesis}	
\textbf{CT from MR} In many clinical settings, the acquisition of CT images is required. 
However, CT imaging puts the patient at the risk of cell damage and cancer because of radiation exposure.
This motivates the synthesis of CT images from MR acquisitions. 
\cite{nie2017medical} synthesize CT images from corresponding MR images with the help of a cascade of 3D Fully Convolutional Networks. In order to improve realism of the synthetic CT images they train the model with a pixel-wise reconstruction loss and an image gradient loss in addition to the adversarial training. 
The idea of utilizing a cascade of generators originates from the so-called Auto-Context Model (ACM). In ACMs a network provides its output as additional input to a succeeding network to provide context information and allow for refinements (Figure \ref{fig:syn-nie2017}). 
%\begin{figure}
%	\centering
%	\includegraphics[width=0.7\linewidth]{Diagrams/nie2017medical}
%	\caption{Proposed architecture by \cite{quan2017compressed}}
%	\label{fig:syn-nie2017}
%\end{figure}
While \cite{nie2017medical} require corresponding pairs of CT and MR images for training, \cite{wolterink2017deep} successfully utilize cycleGANs to transform 2D MR images to CT images without the need for explicit, co-registered data pairs. 
Interestingly, their training led to even better results as the mapping is not affected by co-registration artifacts.
\cite{zhao2018craniomaxillofacial} use conditional GANs to map 3D MR data of the head to its CT counterpart to facilitate the segmentation of craniomaxillofacial bony structures. 
To obtain feasible image-to-image translation results, they propose so-called ``deep supervision discrimination'', which---similarly to the perceptual loss---utilizes the feature representations of a pretrained VGG16 model to i) tell real and synthetic CT images apart and ii) provide gradient updates to the actual generator network.

\textbf{MR from CT} Similar to \citep{wolterink2017deep}, \cite{chartsias2017adversarial} successfully leverage cycleGANs for unpaired image-to-image translation to generate pairs of cardiac MR image and its corresponding segmentation mask from pairs of cardiac CT slices and its ground-truth segmentation. The authors have shown that the performance of a segmentation model can be improved by 16\% when the model is additionally trained with the synthetic data.
They also showed that a model trained with only synthetic data performs only 5\% worse than a model trained on real data. 
A major limitation of GANs, as pointed out by \cite{cohen2018distribution}, is the lack of a guarantee that tumors$/$lesions are preserved during image-to-image translation . 
To cope with this issue, \cite{jiang2018tumor} propose a tumor-aware loss function for cycleGAN to synthesize MR images from CT images.

\textbf{Retinal Image Synthesis}  \cite{costa2017towards} utilize a slight modification of the Pix2Pix framework \citep{Isola2017ImagetoImageTW} for the generation of high resolution eye fundus image from binary images of vessel trees. 
In follow-up work \cite{costa2017end} further introduce an adversarial autoencoder (AAE) which is trained to compress vessel tree images into a multivariate normal distribution and also consecutively reconstructing them. 
The resulting model synthesizes arbitrary high resolution vessel tree images by sampling from the multivariate normal distribution. 
The synthetic vessel tree images in turn can be fed into the image-to-image translation model, leading to an end-to-end framework for realistic, high resolution retinal image synthesis. 
% PROBLEM: The authors do not mention why they use the Adversarial autoencoder to ensure global regularity of the generated vessel trees - a problem with GANs
Very similarly, \cite{guibas2017synthetic} propose a two-stage approach, consisting of a GAN trained to synthesize vessel tree images from noise, and a Pix2Pix network \citep{Isola2017ImagetoImageTW} to generate realistic, high resolution pairs of groundtruth vessel segmentation and corresponding eye fundus image. 
In succession, the authors investigate the performance of a U-Net trained for segmentation using real data pairs and another model trained only on the synthetic samples, and find that training from only the synthetic data leads to a slightly inferior model.

\cite{zhao2017synthesizing} also leverage the Pix2Pix framework for synthesizing filamentary structured images, i.e. eye fundus images and neurons from binary segmentation masks. 
In comparison to \cite{costa2017towards,costa2017end}, the authors also provide their framework with a reference image for style.
To enable style transfer, they train the generator with the feedback of a VGG-network in addition to the discriminator. 
Opposed to Pix2Pix, they do not introduce noise with the help of dropout, but by augmenting noise to the bottleneck of the encoder-decoder network. 

% PROBLEM: THey do not elaborate thoroughly on the high resolution neuron images, leaving unanswered whether high res image synthesis really works here, there are not quantitative results on neurons provided

\textbf{PET from CT} PET images are frequently used for diagnosis and staging in oncology.
The combined acquisition of PET and anatomical CT images is a standard procedure in clinical routine. Furthermore, PET/CT imaging is becoming an important evaluation tool for new drug therapies. However, PET devices are expensive and involve radioactivity, thus put patients at risk. Consequently, the medical imaging analysis community has been working on synthesizing PET images directly from CT data.
Initial promising results to synthesize liver PET images from CT data with a conditional GAN have been obtained by \cite{ben2017virtual}. 
The cGAN, inspired by \cite{Isola2017ImagetoImageTW}, is able to synthesize very realistic looking PET images.
However, its performance has low response to underrepresented tumor regions, which leads to poor tumor detection performance in a set of use-case experiments. 
In comparison, an FCN for PET image synthesis is capable of synthesizing tumors, but produces blurry images in general. 
By blending corresponding synthetic PET images coming from the conditional GAN and the FCN, they are able to achieve very high tumor detection performance. 
% PROBLEM: They notice that FCN produces better malignant tumors, but the images are more blurry, and the cGAN produces more realistic images in general. But they do not make any statement on why the cGAN has problems with the tumor.
Similarly, \cite{bi2017synthesis} utilize a cGAN for synthesizing high-resolution PET images from pairs of CT images and binary labelmaps. 
While CT images alone would be sufficient as input, the authors emphasize that adding a labelmap leads to globally more realistic, synthetic output. 
Because of the two-channel input to the generator, they refer to their network as the multi-channel GAN. 
Further, the authors validated their synthetic PET images with a tumor detection model trained on synthetic data and obtained comparable results to a model trained with real data 
They show that synthetic data can be beneficial when labeled data is scarce.
% PROBLEM: But why didnt they augment the data? What was wrong?

\textbf{PET from MRI} Measuring the myelin content in PET images of the human brain is valuable for monitoring disease progression, understanding physiopathology and evaluating treatment efficacy of Multiple Sclerosis (MS). 
Unfortunately, PET imaging for MS is costly and invasive as it requires the injection of a radioactive tracer. 
\cite{wei2018learning} successfully utilize a cascade of two conditional GANs, based on 3D U-Nets for the generators and 3D CNNs for the discriminators, to synthesize PET images from a set of different MR volumes. 
The authors noted that a single cGAN was insufficient for the task at hand as it produced blurry images. Splitting the synthesis task into smaller, more stable subproblems seemed to drastically improve the results.

\textbf{Ultrasound} \cite{hu2017freehand} propose a cGAN conditioned on 3D voxel locations for synthesizing 2D freehand Ultrasound (US) images of a fetus phantom. 
In comparison to the standard cGAN, the authors find it necessary to transform the pixel locations into featuremaps and inject these at various layers of the generator to facilitate training. 
In their experiments, they demonstrate the capability of simulating US images at locations which are unseen to the network. Further, they quantify the generation of sound images by comparing the location of clinically relevant anatomical landmarks in synthetic and real images, and verify the realism of the generated images in a usability study.

% PROBLEM: Blurring is kinda cheating, right?
\cite{tom2017simulating} apply GANs for intravascular US (IVUS) simulation in a multi-stage setup. 
A first generator, conditioned on physically simulated tissue maps, produces speckle images. 
The second generator maps the speckle images to low resolution, synthetic $64\times 64$ pixels sized US images. 
A third generator transforms these low resolution images into high resolution samples at a resolution of $256\times 256$ pixels. 
In a visual turing test, the synthetic images could not reliably be distinguished from real ones.

\textbf{X-ray} \cite{mahapatra2018efficient} train a cGAN for chest X-ray synthesis from perturbed segmentation maps and combine it with Bayesian Neural Networks for Active Learning. 

\textbf{Stain Normalization} Conditional GANs have also been leveraged for coping with the variance in digital histopathology staining, which is well known to cause problems for CAD systems. 
\cite{cho2017neural} point out that tumor classifiers do not only generalize poorly across data with different stains, but also that existing stain-normalization approaches fail to preserve important image features. 
To overcome this issue, they propose a feature-preserving cGAN for stain style transfer.
First, they map histopathology images to a canonical gray-scale representation. 
In succession, they leverage a cGAN to transform these gray-scale images into RGB images with the desired staining. 
The benefits of employing an additional feature-preserving loss on the hidden layers of the discriminator are experimentally validated by obtaining the smallest performance degradation among a variety of State-of-the-Art (SOTA) stain transfer methods.

\cite{bayramoglu2017towards} leverage the Pix2Pix framework for virtual H\&E staining on unstained hyperspectral microscopy patches. 
The authors report promising quantitative results, but also point out to require expert feedback in order to draw a valid conclusion.

\cite{bentaieb2018adversarial} try to tackle the stain transfer problem with the help of an AC-GAN by simultaneously training a conditional GAN for stain-transfer and a task-specific network (i.e. a segmentation or classification model). 
The joint optimization of the generator, the discriminator and the task-specific network drives the generator to produce images with relevant features being preserved for the task-specific model.

Aformentioned methods rely on paired training data to map from a source to target staining.
Such a dataset is often hardly available and requires preprocessing such as co-registration. 
Moreover, co-registration itself is not perfect and is prone to produce artifacts. \cite{shaban2018staingan} alleviate the need for paired training data and co-registration by employing cycleGANs for the task of stain transfer. 
In a broad set of experiments on different datasets, they obtain visually more compelling stain transfer results than previous deep-learning and non-deep learning based methods.

\textbf{Microscopy} \cite{han2017transferring} propose a framework similar to Pix2Pix for transferring Phase Contrast (PC) and Differential Interference Contrast (DIC) in microscopy images to one another. 
A U-net like generator is trained to synthesize the image of a certain modality from a source modality and the corresponding cell mask. 
Interestingly, two discriminators are employed to differentiate between i) pairs of (real source, real target) versus (real source, synthesized target), and ii) pairs of (cell mask real source) versus (cell mask, synthesized target). 
In a set of qualitative and quantitative evaluations they rank their two-discriminator approach against the single-discriminator Pix2Pix framework. They report improved quantitative results when transferring from DIC image to Phase Contrast, and comparable results when trying to map from Phase Contrast to DIC. The authors amount the comparable performance of the latter to the details already present in PC images, which leaves the cell mask with very little impact on the synthesis outcome.

\textbf{Blood Vessels} Machine Learning driven analysis methods for detecting atherosclerotic plaque or stenosis in coronary CT angiography (CCTA) are powerful, but data-hungry. 
To deal with the lack of labeled data, \cite{wolterink2018blood} propose to synthesize plausible 3D blood vessel shapes with the help of a WGAN fed with noise and attribute vectors. 
To facilitate the synthesis in 3D at appropriately high resolution, the authors generate 1D parameterizations of primitives which characterize blood vessels as a proxy for the final vessel rendering.
Alternatively to CT, Magnetic Resonance Angiography (MRA) has evolved as an important tool for visualizing vascular structures
However, often it is not acquired alongside the standard protocols. 
\cite{olut2018generative} propose the so-called steerable GAN to synthesize MRA images from T1 and T2-weighted MR scans, potentially alleviating the need for additional MR scans. 
Their conditional, steerable GAN combines a ResNet-like generator with a PatchGAN-discriminator. 
Also they propose an $\ell_1$-loss between real and synthesized image as well as a steerable filter loss to promote faithful reconstructions of vascular structures.

%Conc

The unconditional methods are summarized in Table \ref{table:unconditional_synthesis}, whereas the conditional GAN variants are summarized in Table \ref{table:conditional_synthesis1} and \ref{table:conditional_synthesis2}. 
In particular, we report the method, containing the underlying GAN architecture, the image modalities on which the particular method operates, the datasets which have been used and the resolution of the synthesized images. 
We further mark whether a paper has been peer-reviewed or not. 
Since losses are a substantial part of the underlying GAN framework, we do not explicitly report them here. 
Further, we do not report any quantitative results since they i) are in many case unavailable, ii) hardly interpretable and iii) overall hardly comparable. 

In general, many interesting GAN-based approaches have been proposed for both unconditional and conditional image synthesis. 
However, often the validity of the method at hand is questionable and requires more elaboration. 
For instance, in many visual turing tests it seemed to have been fairly easy to distinguish between real and generated images \citep{frid2018synthetic,chuquicusma2017fool,hu2017freehand} due to artifacts in synthetic samples. 
\cite{hu2017freehand} and \cite{chuquicusma2017fool} tackle this problem by applying anisotropic or gaussian filtering to both real and fake samples before presenting them to the raters \citep{hu2017freehand, chuquicusma2017fool}. 
This is only valid as long as blurry images still contain the required amount of information for the task at hand. 
Another problem of GAN is mode collapse, resulting in very similar looking synthetic samples. 
Particularly in the works of \cite{kitchen2017deep} and \cite{schlegl2017unsupervised}, where samples look fairly similar, it should be considered if mode collapse has occurred. 
In general, the community still lacks a meaningful, universal quantitative measure for judging realism of synthetic images. 
Regardless of the realism, aforementioned works have shown that GANs can be used successfully for data simulation and augmentation in classification and segmentation tasks. 
How exactly realism, artifacts and other specific properties of generated samples affect a machine learning model generally remains an open question.

\begin{table}[!ht]
	\tiny
	\centering\renewcommand\cellalign{lc}
	\caption{Unconditional GANs for Medical Image Synthesis}
	\label{table:unconditional_synthesis}
	\begin{adjustbox}{width=\textwidth}{%
			\begin{tabu} to \textwidth {@{}X[0.7l] X[0.4l] X[0.4l] X[0.8l] X[0.3l] X[0.1l] X[0.15l]@{}}
				\toprule
				Method 	    &		Architecture		& 		Modality 		& 		Dataset 		&      Resolution		& 	PR   & Code \\ \hline
				
				\textbf{\cite{kitchen2017deep}} &  	
				DCGAN &		
				\makecell{MRI \\Prostate \\Lesions}  & 
				\makecell{SPIE ProstateX \\Challenge 2016\\(330 MRI Scans)}  & 
				16$\times$16 &
				No &
				No \\ \hline
				
				%\makecell{\textbf{\cite{schlegl2017unsupervised}}\\\textbf{Architecture}:\\DCGAN} 		&  			
				%\makecell{Retinal OCT}  & 
				%non-public  & 
				%64$\times$64 \\ \hline
				
				\textbf{\cite{chuquicusma2017fool}} 		&
				DCGAN &  			
				\makecell{CT Lung \\Cancer Nodules}  & 
				\makecell{LIDC-IDRI\\(1145 nodules)} & 
				56$\times$56 &
				Yes &
				No \\ \hline
				
				%\makecell{\textbf{\cite{basusynthesis}}\\\textbf{Architecture}:\\Wasserstein-GAN} 		&  			
				%\makecell{GAD-Enhanced 2D brain MR}  & 
				%NeuroRx. MS-LAQ  & 
				%64$\times$64 \\ \hline
				
				\textbf{\cite{frid2018synthetic}}		&  
				DCGAN &			
				\makecell{focal CT liver \\lesion patches}  & 
				\makecell{non-public\\(182 CT scans)}  & 
				64$\times$64 &
				Yes &
				No \\ \hline
				
				\textbf{\cite{bermudez2018learning}}		&  			
				DCGAN &
				\makecell{2D axial brain\\MR slices}  & 
				\makecell{Baltimore Longitudinal\\Study of Aging (BLSA)\\(528 healthy subjects)}  & 
				220$\times$172 &
				Yes &
				No \\ \hline
				
				\textbf{\cite{baur2018melanogans,baur2018generating}} 		&

				DCGAN, LAPGAN, PGAN &  			
				\makecell{Skin Lesions\\(Dermoscopy)}  & 
				\makecell{ISIC2017 (2000 samples)\\ISIC2018 (10,000 samples)}  & 
				256$\times$256 &
				No & %Workshop
				No \\ \bottomrule
			\end{tabu}%
		}
	\end{adjustbox}
\end{table}

\begin{table}[!ht]
	\tiny
	\centering\renewcommand\cellalign{lc}
	\caption{Conditional GANs for Medical Image Synthesis}
	\label{table:conditional_synthesis1}
	\begin{adjustbox}{width=\textwidth}{%
			\begin{tabu} to \textwidth {@{}X[0.65l] X[0.95l] X[l] X[1.3l] X[0.5l] X[0.1l] X[0.3l]@{}}
				\toprule
				Method 	    &		Architecture		& 		Modality 		& 		Dataset 		&    Resolution		&		PR  & Code \\ \hline		
				\textbf{\cite{nie2017medical}} &	
				\makecell{3D Autocontext\\FCN with adversarial\\loss, image gradient\\loss and $\ell_2$-loss} 		&  		
				\makecell{MR to CT}  & 

				\makecell{ADNI (16 subjects)\\ and 22 non-public\\pelvic image pairs} & 
				32$\times$32$\times$32 (MRI) to 16$\times$16$\times$16 (CT) &
				Yes &
				Yes \\ \hline
				
				% \cite{wei2018learning}: PET to MRI for myelin
				
				\textbf{\cite{wolterink2017deep}} &
				cycleGAN 	&  			
				\makecell{2D saggital brain MR\\ and CT slices}  & 
				non-public (24 subjects) & 
				256$\times$256 &
				No & %Workshop
				No \\ \hline
				
				\textbf{\cite{zhao2018craniomaxillofacial}} &
				3D cond. GAN 		&  			
				\makecell{MR to CT}  & 
				\makecell{ADNI (16 subjects)} & 
				152 $\times$ 184 $\times$ 149 &
				Yes &
				No \\ \hline
				
				\textbf{\cite{chartsias2017adversarial}} & 
				cycleGAN 		&  			
				\makecell{2D cardiac MR w. \\segmentation mask to \\cardiac CT w. \\segmentation mask}  & 
				\makecell{non-public\\(20 CT/CTA and 20 MRI)} & 
				232$\times$232 &
				No & %Workshop
				No \\ \hline
				
				\textbf{\cite{jiang2018tumor}} &
				Tumor-Aware cycleGAN 		&  			
				\makecell{CT to MR}  & 
				\makecell{NSCLC (The Cancer\\Imaging Archive, 377 scans)\\\& non-public (42 scans)} & 
				256$\times$256 &
				Yes &
				No \\ \hline
				
				\textbf{\cite{costa2017towards,costa2017end}} &
				AAE and Pix2Pix 		&  			

				\makecell{2D binary vessel tree\\images to retinal\\ images}  & 
				\makecell{DRIVE (40 samples)\\MESSIDOR (1200 samples)} & 
				512$\times$512 &
				Yes &
				Yes\\ \hline
				
				\textbf{\cite{jin2018ct}} &
				3D cond. GAN 		&  			
				\makecell{CT (lung nodules)}  & 
				\makecell{LIDC (1018 scans)} & 
				64$\times$64$\times$64 &
				Yes &
				No \\ \hline
				
				\textbf{\cite{guibas2017synthetic}} &
				GAN and Pix2Pix 		&  			
				\makecell{2D binary \\vessel tree,\\images to retinal}  & 
				\makecell{DRIVE (40 samples)\\MESSIDOR (1200 samples)} & 
				512$\times$512 &
				No &
				Yes \\ \hline
				
				\textbf{\cite{zhao2017synthesizing}} &
				Pix2Pix w. Style Transfer 		& 
				\makecell{eye fundus,\\microscopic neuronal}  & 
				\makecell{DRIVE (40 samples)\\STARE (20 samples)\\HRF (45 samples)\\NeuB1 (112 samples)} & 
				512$\times$512 and higher &
				No &
				Yes \\ \hline
				
				\textbf{\cite{ben2017virtual}} &
				Pix2Pix and FCN 		& 
				\makecell{2D liver tumor \\CT to PET}  & 
				\makecell{non-public (25 pairs)} & 
				n/a &
				No & %Workshop
				No \\ %\hline
				\bottomrule
			\end{tabu}%
		}
	\end{adjustbox}
\end{table}

\begin{table}[!ht]
	\tiny
	\centering\renewcommand\cellalign{lc}
	\caption{Conditional GANs for Medical Image Synthesis}
	\label{table:conditional_synthesis2}
	\begin{adjustbox}{width=\textwidth}{%
			\begin{tabu} to \textwidth {@{}X[0.65l] X[0.95l] X[l] X[1.3l] X[0.5l] X[0.1l] X[0.3l]@{}}
				\toprule
				Method 	    &		Architecture		& 		Modality 		& 		Dataset 		&    Resolution		&		PR & Code \\ \hline	
				
				\textbf{\cite{bi2017synthesis}} &
				cond. multi-channel GAN 		& 
				\makecell{CT and segmentation\\pairs to PET images}  & 
				\makecell{non-public \\(50 subjects)} & 
				200$\times$200 &
				No & %Workshop
				No \\ \hline
				
				\textbf{\cite{hu2017freehand}} &
				spatially cond. GAN 		& 
				\makecell{2D US}  & 
				\makecell{non-public fetus phantom\\(26,396 images)} & 
				160$\times$120 &
				No & %Workshop
				Yes \\ \hline
				
				\textbf{\cite{tom2017simulating}} &
				multi-stage cond. GAN 		& 
				\makecell{simulated tissue\\maps to 2D\\Intravascular US}  & 
				\makecell{IVUS challenge\\(2,175 images)} & 
				256$\times$256 &
				Yes &
				No \\ \hline
				
				\textbf{\cite{mahapatra2018efficient}} &
				cond. GAN 		& 
				\makecell{Segmentation maps to\\synthetic X-ray images}  & 
				\makecell{SCR chest XRay database\\(247 images)\\NIH (400 images)} & 
				512$\times$512 &
				Yes &
				No \\ \hline
				
				\textbf{\cite{cho2017neural}} &
				feature-preserving\\cond. style-transfer GAN 		& 
				\makecell{Digital\\Histopathology}  & 
				\makecell{CAMELYON16\\(400 slides)} & 
				n/a &
				No &
				No\\ \hline
				
				\textbf{\cite{bayramoglu2017towards}} &
				Pix2Pix 		& 
				\makecell{Hyperspectral\\ microscopic \\to H\&E stained}  & 
				\makecell{non-public\\(2838 image pairs)} & 
				64$\times$64 &
				No & %Workshop
				Yes \\ \hline
				
				\textbf{\cite{bentaieb2018adversarial}} &
				ACGAN 		& 
				\makecell{Digital\\Histopathology}  & 
				\makecell{MITOS-ATYPIA14\\(11 slides)\\MICCAI’16 GlaS challenge\\(165 slides)\\non-public ovarian\\carcinoma (135 slides)} & 
				250$\times$250 &
				Yes &
				No \\ \hline
				
				\textbf{\cite{shaban2018staingan}} &
				cycleGAN 		& 
				\makecell{Digital\\Histopathology}  & 
				\makecell{MITOS-ATYPIA14\\(11 slides)\\Camelyon16\\(400 slides)} & 
				256$\times$256 &
				Yes &
				Yes \\ \hline
				
				\textbf{\cite{han2017transferring}} &
				cond. GAN with two Discriminators 		& 
				\makecell{DIC \& Phase Contrast \\Microscopy}  & 
				\makecell{non-public\\(1,600 pairs)} & 
				256$\times$256 &
				No & %Workshop
				No \\ \hline
				
				%\makecell{\textbf{\cite{Chen2018EfficientAA}}\\\textbf{Architecture}:\\3D CNN with\\adversarial training} 		& 
				%\makecell{T1w MRI Superresolution}  & 
				%The human connectome project &
				%320$\times$320$\times$256 \\ \hline
				
				%\makecell{\textbf{\cite{sanchez2018brain}}\\\textbf{Architecture}:\\3D SRGAN} 		& 
				%\makecell{T1w MRI Superresolution}  & 
				%ADNI & 
				%128$\times$128$\times$92 patches \\ \hline
				
				%\makecell{\textbf{\cite{ravi2018adversarial}}\\\textbf{Architecture}:\\CycleGAN} 		& 
				%\makecell{Endomicroscopy (pCLA) Superresolution}  & 
				%non-public & 
				%64$\times$64 patches \\ \hline
				
				\textbf{\cite{wolterink2018blood}} &
				WassersteinGAN 		& 
				\makecell{Geometric parameters \\extracted from CCTA}  & 
				\makecell{non-public\\(4,412 centerlines)} & 
				n/a &
				No &
				No \\ \hline
				
				\textbf{\cite{olut2018generative}} &
				cond. steerable GAN 		& 
				\makecell{MRA from T1 \& \\T2w MRI \\axial slices}  & 
				\makecell{IXI Dataset (578 pairs)} & 
				n/a &
				No & %Workshop
				No \\ \bottomrule
			\end{tabu}%
		}
	\end{adjustbox}
\end{table}

%% file: sections/applications/segmentation.tex
\subsection{Segmentation}
\label{sub:Segmentation}

%introduction
Segmentation of objects and organs in medical images is an essential pre-requisite for many applications such as detection, classification, and shape analysis.
The tedious and time-consuming nature of manual segmentation made automatic methods the most active field of Deep-Learning research in medical image analysis \citep{litjens2017survey}.
However, the pixel-wise based evaluation and optimization procedure in deep networks is not sufficient to extract notions of anatomical structures.
This drawback requires additional corrections on top of CNNs such as Conditional Random Fields (CRFs) and Statistical Shape Models (SSMs), which are not easily optimizable \citep{chen2016deeplab,tack2018knee}.
GANs, providing a different learning flow, are a potential solution to address this problem. 
In the following Subsection we focus on GAN-based segmentation approaches, categorized based on specific parts of the anatomy.

\subsubsection{Brain:}
\label{app:seg:brain}
%intro
The proposed methods for brain structures and abnormality segmentation comprise both supervised and unsupervised methods. 
%GAN
\cite{moeskops2017adversarial} demonstrate that using the GAN training strategy in CNNs enhances not only the performance of semantic segmentation methods but also brings the performance of non-semantic segmentation methods closer to semantic ones.
\cite{li2017brain} highlight the superior performance of GANs in the segmentation of normalized/equalized patches of brain tumors.
\cite{Xue2018SegANAN} propose the SegAN framework which employs the U-Net as the generator architecture of GAN. 
They show that pixel-dependencies are learned better by using a multi-scale loss function in addition to adversarial and pixel-wise losses.     		
%unsupervised GAN
One of the known challenges with most of supervised segmentation methods is the performance degradation on unseen images. 
\cite{kamnitsas2017unsupervised} propose to leverage an adversarial framework to address this problem for unsupervised domain adaptation in brain lesion segmentation. 
In this method, a domain discriminator network is employed to make the segmentation network invariant to input from different domains.
% with the help of GAN in the synthesis	
The varying appearance of anatomical tissues in multi-modal images leads to improved segmentation performance.
For the segmentation of bony structures in brain images \cite{Zhao2018CraniomaxillofacialBS} propose to synthesize CT images from MRI images using GANs and then use both of them as the input to a segmentation network.
They proposed a segmentation architecture called Deep-supGAN which is optimized with 5 different losses: an adversarial loss, a voxel-wise loss, and 3 perceptual losses defined on differences between VGG extracted features.
%cGAN, MGAN
For multi-class classification of brain tumors, \cite{Rezaei2017ACA} propose a method combining the cGAN and MGAN, where class labels define conditions.    	

\subsubsection{Chest:}
\label{app:seg:chest}	
%Intro
Bad quality, local artifacts and the overlap of lung and heart area are the main obstacles for segmentation of chest X-Ray images. 
As existing approaches in this field do not provide consistency between global and local features, \cite{dai2018scan} propose the SCAN architecture, achieving human levels of performance in heart/lung segmentation.
In this method, the generator is pre-trained with a pixel-wise loss function to address the instability problem of GANs.

\subsubsection{Eye:}
\label{app:seg:Retina}
%Intro
Many CNN-based approaches perform even better than human experts in retinal vessel segmentation. 
However, blurriness and false positive segmentations near minuscule or faint branches constitute a problem which is not solved yet.
%GAN
\cite{son2017retinal,lahiri2017generative} propose GANs as a solution to this problem.
%DCGAN
%\cite{lahiri2017generative} propose a patch based segmentation using DC-GAN. 
%cGAN
For optic disc and cup segmentation in 2D color fundus images, \cite{shankaranarayana2017joint} successfully leverage a cGAN.

\subsubsection{Abdomen:}
\label{app:seg:Abdomen}
%GAN
\cite{yang2017automatic} propose GANs which employ the U-Net as the generator to segment liver in 3D CT images of the abdomen.
%cycleGAN
\cite{kim2018cycle} propose to use the cycleGAN for both liver and liver tumor segmentation. 
To address the problem of miss-segmentation of tiny tumors, they propose a new architecture based on the U-Net (polyphase U-Net) for the generator.
%cGAN
Spleen segmentation in MRI images is a challenging task due to the varying size and shape of this organ.
To address this problem \cite{huo2017splenomegaly} propose a cGAN combined with the global convolutional network (GCN) \citep{Peng2017LargeKM} as the generator architecture. 
They showed that larger convolutional kernels in the GCN in addition to adversarial training enhances the segmentation performance on objects with large variability. 

\subsubsection{Microscopic images:}
\label{app:seg:microscopic}
Automatic segmentation of microscopic images is challenging due to the variety of size, shape, and texture. \citep{Sadanandan2017SpheroidSU,Arbelle2018MicroscopyCS}. 
%GAN
\cite{Sadanandan2017SpheroidSU} propose to use GANs with special training loss function, which considers a weight to specify which pixels in foreground/background are more important.
\cite{Arbelle2018MicroscopyCS} use GANs with special blocks (convolution followed by batch normalization) in the discriminator for the same problem. 
%DCGAN
\cite{Zhang2017DeepAN} propose DAN -- a combination of DCAN \citep{Chen2016DCANDC} and VGG16 -- which is trained with both supervised and unsupervised strategies to provide high-quality segmentation masks for unseen images. 

\subsubsection{Cardiology:}
\label{app:seg:Cardic}
%Intro
Low contrast, high level of noise, and cardiac motion are challenges for segmenting cardiology images.
%cGAN
To segment the Left Ventricle (LV) in low-contrast cardiology images, \cite{Dong2018VoxelAtlasGAN3L} propose the VoxelAtlasGAN which employs a V-Net \citep{Milletari2016VNetFC} atlas-based segmentation in the generator of a cGAN framework (Figure \ref{fig:seg1}). 
Also, \cite{Xu2018MuTGANSS} propose to benefit from the cGAN on top of atlas-based segmentation to facilitate feature extraction of all time sequences frames for precise segmentation of myocardial infarction.
%GAN	
Since in segmentation, residual (non-RoI) information in addition to RoI features can make segmentation results more realistic, \cite{Chartsias2018FactorisedSR} and \cite{Joyce2018DeepMS} propose a reconstruction strategy based on both of these features. They added a reconstruction loss to the optimization procedure of GAN as a controlling parameter.
%cycleGAN and LSGAN
For semi-supervised and unsupervised segmentation, \cite{Chartsias2018FactorisedSR} and \cite{Joyce2018DeepMS} propose to use the cycleGAN and LSGAN architectures, respectively. 
They also proposed some loss functions to address over-segmentation in large regions with varying intensity. 
%\begin{figure}
%	\centering
%	\includegraphics[width=0.7\linewidth]{Diagrams/Dong2018VoxelAtlasGAN3L}
%	\caption{Proposed architecture by \cite{Dong2018VoxelAtlasGAN3L}}
%	\label{fig:seg1}
%\end{figure}

\subsubsection{Spine:}
\label{app:seg:spine}
%Intro
In medical image analysis of the spine, machine learning based approaches suffer from not properly learning the anatomy of the discs in vertebrae for segmentation and localization. 
%GAN
\cite{Sekuboyina2018BtrflyNV} propose a butterfly shape GAN model to segment disc regions in two views of vertebrae.

%Conclusion
\Cref{table:seg:brain,table:seg:chest,table:seg:eye,table:seg:Abdomen,table:seg:Micr,table:seg:Cardi,table:seg:spine} summarizes GAN-based segmentation methods. These mainly use GAN/cGAN/DCGAN in addition to pixel/voxel-wise optimization loss functions. 
Also, CycleGAN and reconstruction loss strategies are proposed to consider non-RoI features for more precise segmentation.
In the reviewed methods, U-Net and ResNet - due to providing general identification features - are the most popular segmentation networks for the generator architecture. 
However, in the reviewed papers, some limitations exist which trouble a clear judgment on the proposed methods.
%benefits
The ability of GANs to consider both global and local information performs close to semantic segmentation, even though difficulties related to semantic methods, unsupervised learning, and detail preservation are addressed. 
Tables show that GAN-based methods are trainable using a varying amount of data from 10 to 1000+ input samples.
%limitation
However, there are some limitations in the information provided in the reviewed papers.
Some papers only used the DICE similarity score for evaluation, which its practical meaning is debatable.

Experimental results generally show that the competition between segmentors using and not using adversarial training is tightly closed. 
However in microscopic images segmentation, GANs made a significant enhancement.

\begin{table}[!ht]
	\tiny
	\centering\renewcommand\cellalign{lc}
	\caption{Segmentation GAN-based methods - Brain}
	\label{table:seg:brain}
	\begin{adjustbox}{width=\textwidth}{%
		\begin{tabu} to \textwidth {@{}X[0.8l] X[0.4l] X[0.53l] X[0.3l] X[0.6l] X[1.4l] X[0.1l] X[0.2l]@{}}
			\toprule
			method 	& Arch & loss &	modality & dataset & performance & PR & code \\ \hline
			
			\cite{moeskops2017adversarial} & GAN & \makecell{Adv, \\cross entropy} &  MRI & \makecell{MICCAI 2012 \\Challenge \\Train:15 \\Test:20 \\MRBrainS13 \\ challenge\\Train:5 \\Test:15 } & \makecell{DSC = 0.92$\pm$0.03\\\\\\\\DSC = 0.85$\pm$0.01\\\\\\\\} &Yes &No   
			\\ 
			\hline
			
			\cite{li2017brain} & GAN & Adv &  MRI & \makecell{BRATS 2017 \\Train:285 \\ }& \makecell{(Whole, Core, Enhancing)\\DSC = 0.87, 0.72, 0.68 \\ sensitivity = 0.87, 0.72, 0.68}  &Yes &No 
			\\
			\hline
			
			\makecell{\cite{Xue2018SegANAN}\\ \textbf{SeGAN}}& \makecell{U-Net,\\ GAN} & \makecell{Adv, \\multiScale} & MRI & \makecell{BRATS 2013 \\Train:25\\\\BRATS 2015  \\Train:274\\} & \makecell{(whole, Core, Enhanced)\\DSC = 0.84, 0.70, 0.65\\Precision = 0.87, 0.80, 0.68\\Sensitivity = 0.83, 0.74, 0.72\\DSC = 0.85, 0.70, 0.66\\Precision = 0.92, 0.80, 0.69\\Sensitivity = 0.80, 0.65, 0.62}  &Yes &No       
			\\ 
			\hline
			
			\cite{kamnitsas2017unsupervised} & \makecell{GAN, \\3D-CNN} & \makecell{Adv,\\ SGD} &  MRI (TBI)  & \makecell{unknown\\Train:61 \\} & \makecell{DSC = 0.62\\Recall = 0.58\\Precision = 0.71} &Yes &Yes 
			\\ 
			\hline
			
			\makecell{\cite{Zhao2018CraniomaxillofacialBS}\\\textbf{Deep-supGAN}} & \makecell{GAN, \\VGG16} & \makecell{Adv, \\Preceptual, \\voxel-wise} &  MRI 3D  & \makecell{ADNI\\Train:16 \\ } & DSC = 94.46 &Yes &No 
			\\ 
			\hline
			
			\cite{Rezaei2017ACA} & \makecell{c-GAN,\\MGAN } & Adv &  MRI & \makecell{BRATS 2017 \\Train:285 } & \makecell{(Whole, core, Enhanced)\\DSC = 0.70, 0.55, 0.40\\Sensitivity = 0.68, 0.52, 0.99\\Specificity = 0.99, 0.99, 0.99} &Yes &No  
			\\ 
			
			\bottomrule                                                                
		\end{tabu}%
		}
	\end{adjustbox}
\end{table}

\begin{table}[!ht]
	\tiny
	\centering\renewcommand\cellalign{lc}
	\caption{Segmentation GAN-based methods - Chest}
	\label{table:seg:chest}
	\begin{adjustbox}{width=\textwidth}{%
		\begin{tabu} to \textwidth {@{}X[1.4l] X[0.5l] X[1.4l] X[0.7l] X[1.5l] X[1.9l] X[0.5l] X[0.5l]@{}}
			\toprule
			method 	& Arch & loss &	modality & dataset & performance & PR & code \\ \hline	
			\makecell{\cite{dai2018scan}\\ \textbf{SCAN}} & \makecell{VGG,\\ ResNet,\\ GAN} & \makecell{multi-class \\cross-entropy, \\Adv} & X-Ray & \makecell{JSRT\\Montgomery\\ Train:382 \\} & \makecell{(Lungs, Heart)\\DSC = 0.973, 0.927\\IoU = 0.947, 0.866} &Yes  &No 
			\\ 
			\bottomrule                                                                
		\end{tabu}%
		}
	\end{adjustbox}
\end{table}

\begin{table}[!ht]
	\tiny
	\centering\renewcommand\cellalign{lc}
	\caption{Segmentation GAN-based methods - eye}
	\label{table:seg:eye}
	\begin{adjustbox}{width=\textwidth}{%
		\begin{tabu} to \textwidth {@{}X[1.5l] X[0.6l] X[1.2l] X[0.9l] X[1.5l] X[1.9l] X[0.5l] X[0.5l]@{}}
			\toprule
			method 	& Arch & loss &	modality & dataset & performance & PR & code \\ \hline			
			\cite{son2017retinal} & \makecell{U-Net, \\GAN} & \makecell{Adv, \\Cross entropy} & \makecell{Funduscopy \\(Retina)} & \makecell{DERIVE\\Train:20 \\Test:20 \\STARE\\Train:10 \\Test:10 \\}  & \makecell{DSC = 0.829\\ROC = 0.9803\\precision = 0.9149\\DSC = 0.834\\ROC = 0.9838\\precision = 0.9167} &No &Yes \\  
			\hline
			
			\cite{lahiri2017generative} & DCGAN & \makecell{Adv} & \makecell{Funduscopy\\ (Retina)} & \makecell{DERIVE \\(blood vessels)\\Train:20 \\Test:20 }   & AUC = 0.945 &Yes &No
			\\ 
			\hline
			
			\cite{shankaranarayana2017joint} & \makecell{c-GAN, \\ResU-net} & \makecell{Adv,\\$L1$}         & \makecell{Funduscopy\\ (Retina)} & \makecell{RIM-ONE\\Train:159  }    & \makecell{(disc, cup)\\F-score = 0.97, 0.94\\ IOU = 0.89, 0.76} &Yes &No
			\\
			\bottomrule                                                                
		\end{tabu}%
		}
	\end{adjustbox}
\end{table}
			
\begin{table}[!ht]
	\tiny
	\centering\renewcommand\cellalign{lc}
	\caption{Segmentation GAN-based methods - Abdominal}
	\label{table:seg:Abdomen}
	\begin{adjustbox}{width=\textwidth}{%
		\begin{tabu} to \textwidth {@{}X[1.5l] X[1.4l] X[1.3l] X[0.9l] X[1.5l] X[1.9l] X[0.5l] X[0.5l]@{}}
			\toprule
			method 	& Arch & loss &	modality & dataset & performance & PR & code \\ \hline			
			
			\cite{yang2017automatic} & \makecell{U-Net,\\ auto-encoder} & \makecell{Adv, \\cross-entropy}   & \makecell{CT 3D\\(Liver)}  &  \makecell{unknown,\\MICCAI\\(SLiver07)\\Train:1000+ \\} & \makecell{DSC = 0.95\\ASD = 1.90} &Yes &No \\ 
			\hline
			
			\cite{kim2018cycle}& \makecell{U-Net,\\cycleGAN} & \makecell{cycleGAN, \\Cross entropy, \\L2}   & \makecell{CT 3D\\(Liver)}  &  \makecell{LiTS2017\\Train:73 \\Test:9 } & \makecell{(liver, lesion)\\DSC = 0.89, 0.46 \\Recall = 0.94, 0.5\\ Precision = 0.86, 0.48} &Yes &No 
			\\ 
			\hline
			
			\makecell{\cite{huo2017splenomegaly}\\ \textbf{SSNet}} & \makecell{GCN,\\ cGAN} & \makecell{Adv, \\DSC} & MRI & \makecell{unknown\\Train:45 \\Test:15 } & DSC = 0.9260 &Yes &No 
			\\ 
			\bottomrule                                                                
		\end{tabu}%
		}
	\end{adjustbox}
\end{table}

\begin{table}[!ht]
	\tiny
	\centering\renewcommand\cellalign{lc}
	\caption{Segmentation GAN-based methods - Microscopic}
	\label{table:seg:Micr}
	\begin{adjustbox}{width=\textwidth}{%
		\begin{tabu} to \textwidth {@{}X[1.8l] X[0.8l] X[1.3l] X[0.9l] X[1.5l] X[1.9l] X[0.5l] X[0.5l]@{}}
			\toprule
			method 	& Arch & loss &	modality & dataset & performance & PR & code \\ \hline			
			\cite{Sadanandan2017SpheroidSU} & \makecell{GAN, \\U-net, \\Res-Net, \\ MS CNN} & \makecell{Adv, \\weighted loss}   & Cell 2D  &  \makecell{unknown \\Train:5$\times$6000} & \makecell{F-score = 0.70 \\Precision = 0.74 \\Recall = 0.69} &Yes &No \\ 
			\hline
			
			\cite{Arbelle2018MicroscopyCS} & \makecell{GAN \\(rib cage)} & Adv   &  Cell 2D  &   \makecell{H1299\\Train:2 - 11 } & \makecell{F-score = 0.89\\ Precision = 0.82 \\ Recall = 0.85} &Yes &Yes 
			\\ 
			\hline
			
			\makecell{\cite{Zhang2017DeepAN}\\ \textbf{DAN}} & \makecell{GAN, \\DCAN, \\VGG} & \makecell{Adv, \\Multi-scale \\cross entropy}   & \makecell{ Fungus 3D \\}  &   \makecell{2015 MICCAI \\Gland Challenge \\Train:85 \\Test:20} & \makecell{F-score = 0.88\\ DSC = 0.865 \\ OH = 74.55}  &Yes &No 
			\\
		\bottomrule                                                                
		\end{tabu}%
		}
	\end{adjustbox}
\end{table}
			
\begin{table}[!ht]
	\tiny
	\centering\renewcommand\cellalign{lc}
	\caption{Segmentation GAN-based methods - Cardiology}
	\label{table:seg:Cardi}
	\begin{adjustbox}{width=\textwidth}{%
		\begin{tabu} to \textwidth {@{}X[1.8l] X[0.9l] X[1.3l] X[0.9l] X[1.5l] X[1.9l] X[0.2l] X[0.4l]@{}}
			\toprule
			method 	& Arch & loss &	modality & dataset & performance & PR & code \\ \hline			
			\makecell{\cite{Dong2018VoxelAtlasGAN3L}\\ \textbf{VoxelAtlasGAN}} & \makecell{cGAN, \\V-Net} & \makecell{Adv, \\intensity, \\label}   & \makecell{ Echo\\3D}  &   \makecell{unknown\\Train:25 \\Test:35 } & \makecell{DSC = 0.95 \\MSD = 1.85 \\HSD = 7.26 \\corr-of-EF = 0.91 \\time = 0.1} &Yes &No \\
			\hline
			
			\makecell{\cite{Xu2018MuTGANSS}\\ \textbf{MuTGAN}} & \makecell{GAN, \\ConvLSTM, \\3DConv} & \makecell{Adv, \\MAE, \\DSC}   & \makecell{3T MR\\Cardiac\\ cine, \\DE-MR}  &   \makecell{unknown\\Train:140 \\Test: } & \makecell{DSC=0.90 \\Accuracy=96.46 \\Infarct size=22.3} &Yes &No
			\\
			\hline
			
			\cite{Chartsias2018FactorisedSR} & cycleGAN  & \makecell{Adv, \\DSC, \\MAE}   &  \makecell{Cine\\ MR 3D}  &   \makecell{2017 ACDC \\Challenge,\\ Edinburgh \\Imaging \\Facility QMRI\\ Train:128 \\Test:50 } & 
			\makecell{F-score = 0.771} &Yes &Yes 
			\\
			\hline
			
			\makecell{\cite{Joyce2018DeepMS}\\ \textbf{DAN}} & \makecell{LSGAN, \\U-net} &\makecell{Adv, \\Intensity-Var,\\Over-Seg\\ penalty, \\Recons}   & \makecell{ CT or MR\\ Cardiac 2D}  &   \makecell{2017 \\MM-WHS \\Challenge\\Train:20\\ُTest:3MR, 3CT} & \makecell{(MR) DSC = 0.66\\(CT) DSC=0.5}  &Yes &No 
			\\

			\bottomrule                                                                
		\end{tabu}%
		}
	\end{adjustbox}
\end{table}
			
\begin{table}[!ht]
	\tiny
	\centering\renewcommand\cellalign{lc}
	\caption{Segmentation GAN-based methods - Spine}
	\label{table:seg:spine}
	\begin{adjustbox}{width=\textwidth}{%
		\begin{tabu} to \textwidth {@{}X[1.5l] X[0.6l] X[0.6l] X[0.6l] X[1.4l] X[l] X[0.5l] X[0.5l]@{}}
			\toprule
			method 	& Arch & loss &	modality & dataset & performance & PR & code \\ \hline			
			\makecell{\cite{Sekuboyina2018BtrflyNV}\\ \textbf{Btrfly Net}} & \makecell{GAN,\\ Btrfly-Net} & \makecell{Adv, \\Btrfly-Net}   & \makecell{CT 3D}  &   \makecell{\cite{Glocker2013VertebraeLI}\\Train:242 \\Test:60 } & \makecell{Precision=0.84 \\Recall=0.83 \\F1-score= 0.84} &Yes &No 
			\\
			\bottomrule                                                                 
		\end{tabu}%
		}
	\end{adjustbox}
\end{table}

%% file: sections/applications/reconstruction.tex
\subsection{Reconstruction}
\label{sub:reconstruction}

%introduction
Fast MR reconstruction without sacrificing details is a core problem in medical imaging. 
Fast acquisition and reconstruction directly reduces any kind of motion artifacts and is thus highly desirable.

Classic compressed sensing-based solutions directly use k-space information to reconstruct images \citep{yu2017deep}.
The ability to promote realism in images with fast inference makes GANs an obvious candidate for solving the MR reconstruction problem.
GAN-based MR reconstruction research has a major focus on modifying well-known architectures and combining them with appropriate loss functions.

\subsubsection{DAGAN-based strategies}
Early research on GAN-based MR reconstruction focuses on the DAGAN architecture \citep{yang2017dagan}.
In this method, a perceptual loss is added to adversarial and pixel-wise losses to compare deep extracted features in real and generated information, which also enhances the stability of the model.
A series of work refines the DAGAN architecture over time \citep{yu2017deep, yang2017dagan} by manipulating loss functions to preserve frequency information(Figure \ref{fig:recons1}). 
A combined loss function is defined as:   
\begin{equation}
L_{G} = \alpha L_{image-MSE}+\beta L_{freq-MSE}+\gamma L_{VGG}+L_{GAN}
\end{equation}

%\begin{figure}
%	\centering
%	\includegraphics[width=0.7\linewidth]{Diagrams/yang2017dagan}
%	\caption{DAGAN architecture \citep{yang2017dagan}}
%	\label{fig:recons1}
%\end{figure}

The next update of DAGAN is introduced by \cite{seitzer2018adversarial} who propose to add a refinement network (Figure \ref{fig:recons3}) to separate pixel-wise and perceptual information-based training.	

%\begin{figure}
%	\centering
%	\includegraphics[width=0.7\linewidth]{Diagrams/seitzer2018adversarial}
%	\caption{Proposed architecture by \cite{seitzer2018adversarial}}
%	\label{fig:recons3}
%\end{figure}

A similar architecture (Figure \ref{fig:recons2}) is proposed by \cite{quan2017compressed}, where the authors consider a cyclic training strategy based on the reconstruction of data in the lost frequencies.
Moreover, they propose to use a chain of generators to address the ambiguities made in earlier generators. 

%\begin{figure}
%	\centering
%	\includegraphics[width=0.7\linewidth]{Diagrams/quan2017compressed}
%	\caption{RefineGAN architecture \citep{quan2017compressed}: The Generator G is a chain of two concatenated generators (the first generator is for reconstruction and the second one is for refinement) cycle loss is calculated by MSE blocks}
%	\label{fig:recons2}
%\end{figure}

\subsubsection{3D super-resolution strategies}			
\cite{Snchez2018BrainMS} adapt the SRGAN \citep{Ledig2017PhotoRealisticSI} with 3D convolutional layers to deal with volumetric information and enhance the stability of the proposed GAN model. 
Their loss function combines a pixel-wise loss with a Gradient-Based Loss (GDL) \citep{Mathieu2015DeepMV} to address the blurring effect in the reconstructions.
\cite{li2017reconstruction} propose a 3DSRGAN along with two loss functions to control data interpolation and prevent over-fitting.
In addition to the adversarial and the MSE losses, two other objectives are defined to overcome over-fitting and control data interpolation. 
The considerable memory footprint of 3D convolutions is a well-known challenge.
To address this problem, \cite{Chen2018EfficientAA} propose a multi-level Densely Connected Super-Resolution Network (mDCSRN), which is a combination of the WGAN model \citep{Arjovsky2017WassersteinG} and a modified version of DenseNet \citep{Chen2018DeepLabSI}. 

\subsubsection{Other Methods}	
Other GAN-based reconstruction methods mostly introduce additional loss functions to the original framework.
In \citep{mardani2017deep} and \citep{Shitrit2017AcceleratedMR} pixel-wise losses, in \citep{zhang2018multi} and \citep{Han2018ACR} perceptual losses, in \citep{mahapatra2017image} a saliency loss, and in \citep{Rav2018AdversarialTW} Voronoi-vectorization and regularization losses are proposed to be added to the adversarial loss.
\cite{mardani2017deep} proposes to use the LSGAN as a solution to address training instability.

%conclusion	
Table \ref{table:recons:DAGAN}, \ref{table:recons:3D} and \ref{table:recons:others} summarizes properties of mentioned methods and their performance. 
While many architectural modifications of the GAN are proposed, it seems that ResNet is the most popular architecture for the generator.
All the methods introduce a pixel-wise loss in addition to the adversarial loss. 
Also, other loss functions are introduced to preserve essential information in the lost data reconstruction.
%benefits
Due to their synthesis abilities, GANs can provide good performance in the reconstruction of lost data in medical images.
Textural features are essential for diagnosis, and GANs should be trained with additional loss functions to satisfy these aspects. 
Generally, results of reviewed papers show that GANs in comparison with other methods perform faster and more accurate in data reconstruction.
%limitations
Most of the methods need a huge amount of data to provide the convergence through all loss functions defined.
Moreover, evaluation metrics are not necessarily comprehensive to explore the preservation of identical features in the results.

\begin{table}[ht]
	\tiny
	\centering\renewcommand\cellalign{lc}
	\caption{Reconstruction GAN-based methods - DAGAN based}
	\label{table:recons:DAGAN}
	\begin{adjustbox}{width=\textwidth}{%
		\begin{tabu} to \textwidth {@{}X[0.8l] X[0.5l] X[0.53l] X[0.3l] X[0.8l] X[1.4l] X[0.1l] X[0.2l]@{}}
			\toprule
			method 	& Arch & loss &	modality & dataset & performance & PR & code
			\\ \hline
			\cite{yu2017deep} & \makecell{cGAN, \\U-Net} & \makecell{Adv, \\Pix-wise,\\ Perceptual, \\Refinement} 		&  MRI               & \makecell{IXI, MICCAI\\Grand\\Challenge\\2013\\Train:1605+100 \\Test:50 } & \makecell{mask 30\%:\\NMSE=0.09$\pm$0.02\\PSNR=39.53$\pm$4.12\\(CPU, GPU)\\time=0.2$\pm$0.1, 5.4$\pm$0.1(ms)} &No &No           \\ 
			\hline
			
			\makecell{\cite{yang2017dagan}\\ \textbf{DAGAN}}  & \makecell{cGAN,\\U-Net} & \makecell{Adv, \\Pix-wise, \\Frequency,\\Perceptual, \\Refinement}	&  MRI & \makecell{MICCAI\\Grand\\Challenge\\2013\\Train:21128 \\Test:9854} & \makecell{mask 30\%:\\NMSE=0.08$\pm$0.02\\PSNR=40.20$\pm$4.07\\\\(CPU, GPU)\\time= 0.2$\pm$0.1, 5.4$\pm$0.1(ms)}    &Yes &No         \\ 
			\hline
			
			\cite{seitzer2018adversarial} & \makecell{cGAN,\\U-Net} & \makecell{Adv, \\feature \\matching,\\Perceptual, \\penalty}	&  \makecell{MRI \\ Cardic} & \makecell{unknown \\Train:3000 3D \\Test:1200} & \makecell{PSNR=31.82$\pm$2.28\\MOS=3.24$\pm$0.63\\(max=3.78$\pm$0.45)\\SIS=0.94\\(max=1) }  &Yes &No       \\ 
			\hline
			
			\makecell{\cite{quan2017compressed}\\ \textbf{RefineGAN}} & \makecell{GAN chain,\\ ResNet} & \makecell{Adv, \\Cyclic}	& MRI & \makecell{Brain:\\IXI\\Train:100 \\Test:100\\Chest:\\Data Science \\Bowl challenge\\Train:100 \\Test:100}        &  \makecell{mask 30\%, time:0.16(s)\\SSIM=0.97$\pm$0.01\\PSNR=38.71$\pm$2.57\\\\mask 30\%,  time:0.18(s)\\SSIM=0.97$\pm$0.01\\PSNR=38.64$\pm$2.76}  &No &Yes         
			\\ 
			\bottomrule
		\end{tabu}%
		}
	\end{adjustbox}
\end{table}

\begin{table}[!ht]
	\tiny
	\centering\renewcommand\cellalign{lc}
	\caption{Reconstruction GAN-based methods - 3D super-resolution}
	\label{table:recons:3D}
	\begin{adjustbox}{width=\textwidth}{%
		\begin{tabu} to \textwidth {@{}X[0.8l] X[0.6l] X[0.53l] X[0.3l] X[0.8l] X[1.2l] X[0.1l] X[0.2l]@{}}
			\toprule
			method 	& Arch & loss &	modality & dataset & performance & PR & code \\ \hline
			
			\cite{Snchez2018BrainMS} & \makecell{SRGAN, \\subpixel-NN} & \makecell{LSGAN, \\GDL, \\Pixel-wise}	& \makecell{MRI\\(Brain)}	& \makecell{ADNI database\\Train:470\\Test:119}	& \makecell{(Scale 2, Scale 4) \\PSNR=39.28, 33.58\\SSIM=0.98, 0.95} &Yes &No \\ 
			\hline 
			
			\cite{li2017reconstruction} & \makecell{ResNet, \\GAN} & \makecell{Adv, \\Pixel-wise\\3D variation} & \makecell{MRI\\(Brain)} & \makecell{Glioma Patients\\Train:30+10}	& \makecell{PSNR=24.2\\MSE=262.2} &Yes &No \\
			\hline
			
			\makecell{\cite{Chen2018EfficientAA}\\ \textbf{mDCSRN}} & \makecell{DenseNet,\\WGAN} & \makecell{MSE,\\WGAN}	&  \makecell{MRI\\(Brain)} & \makecell{unknown\\Train:891\\Test:111}	& \makecell{PSNR=35.88\\SSIM=0.94\\NRMSE=0.0852} &Yes &No \\
			\bottomrule	
		\end{tabu}%
		}
	\end{adjustbox}
\end{table}

\begin{table}[!ht]
	\tiny
	\centering\renewcommand\cellalign{lc}
	\caption{Reconstruction GAN-based methods - other}
	\label{table:recons:others}
	\begin{adjustbox}{width=\textwidth}{%
		\begin{tabu} to \textwidth {@{}X[0.9l] X[0.9l] X[0.4l] X[0.4l] X[0.8l] X[0.7l] X[0.1l] X[0.2l]@{}}
			\toprule
			method 	& Arch & loss &	modality & dataset & performance & PR & code \\ \hline
			
			\makecell{\cite{mardani2017deep}\\ \textbf{GANCS}\\\\\\\\} & \makecell{ResNet, \\LSGAN}	& Adv & \makecell{MRI\\(Chest)} &  \makecell{contrast-enhanced\\ MRI abdomen \\pediatric \\patients \\Train:300\\Test:50}       &  \makecell{SNR=20.48\\SSIM=0.87\\Time=0.02} &No &No \\ 
			\hline
			
			\cite{Shitrit2017AcceleratedMR} & \makecell{ResNet, \\GAN} & Adv	& \makecell{MRI\\(Brain)}	& \makecell{unknown\\Train:1560 \\Test:346} & PSNR=37.95 &No &No \\ 
			\hline
			
			\makecell{\cite{zhang2018multi}\\ \textbf{GANCS}} & \makecell{ResNet, \\GAN, \\VGG} & Adv	& \makecell{MRI\\(Brain 2D)} &  \makecell{unknown\\Train:170 \\Test:43 }       &  \makecell{PSNR=32.32\\SSIM=0.88\\Time=0.37} &Yes &No \\
			\hline
			
			\cite{mahapatra2017image} & \makecell{ResNet, \\GAN} & \makecell{Adv, \\CNN\\saliency}	&  \makecell{Retinal\\Funduscopy} & \makecell{unknown\\(5000 data)) }	& \makecell{(Scale 4, Scale 8) \\SSIM=0.89, 0.84\\RMSE=6.2, 7.5\\PSNR=44.3, 39db} &Yes &No \\ 
			\hline
			
			\cite{Han2018ACR} & \makecell{\cite{Su2013CellSI}, \\GAN} & \makecell{Adv, \\Perceptual}	&  \makecell{Microscopy \\(Cell)} & \makecell{unknown\\Train:11000 \\Test:500}	& PSNR = 27.8591 &Yes &No \\
			\hline
			
			\cite{Rav2018AdversarialTW} & \makecell{ \cite{ledig2017photo}, \\GAN, \\Cyclic} & \makecell{Adv, \\Regular}	&  \makecell{Endo-\\microscopy} & \makecell{ \cite{andre2011smart}\\Train:202\\Test:36}	& \makecell{SSIM=0.8.7\\$\Delta GCF_{ \widehat{HR} }$=0.66\\$\Delta GCF_{ \widehat{LR} }$=0.37\\$Tot_{cs}$=0.66} &Yes &No \\
			\bottomrule
		\end{tabu}%
		}
	\end{adjustbox}
\end{table}

%% file: sections/applications/detection.tex
\subsection{Detection}
\label{sub:detection}

%introduction
The detection of anomalies from images with supervised Deep Learning algorithms requires a large amount of annotated training data. 
GANs tackle this problem in a different ways: i) by enhancing datasets with synthetic samples, or ii) by modeling distributions from which anomalies can be detected as outliers.

%GANs
For aggressive prostate cancer detection \citep{kohl2017adversarial} and skin lesion detection \citep{udrea2017generative} propose to use the U-Net architecture as the generator of a GAN and a cGAN, respectively.
In prostate US, the low contrast of tumor boundaries challenges contour detection. \citep{Tuysuzoglu2018DeepAC} address this problem using GANs.

%AnoGAN
A different approach towards unsupervised anomaly detection is to model the distribution of normal data with GANs. 
Then the GAN reconstructs the most similar normal image to the query image and anomalies can be detected as discrepancies between the query and the reconstructed image.

\cite{schlegl2017unsupervised}, introducing an unsupervised framework --named AnoGAN, showed that such an idea can be effective in detecting anomalies in OCT images of the retina.
Succeeding work in \citep{Chen2018UnsupervisedDO} and \citep{baur2018deep} modified this framework for anomaly detection in brain MR images. 
Inspired by the AnoGAN, \cite{baumgartner2017visual} propose the Visual Attribution GAN (VA-GAN) for Alzheimer's disease detection.
This model extracts the map of changes that convert the class of the image from healthy to diseased and use it for abnormality detection.
CNNs \citep{shwartz2017opening} show good performance in detecting high contrast lesions arising from Alzheimer's, but the VA-GAN also detects low contrast lesions. 	

%conclution 
Table \ref{table:detect} summarizes these papers. 
Papers proposed in anomaly detection by GANs have more structural complexity in comparison to previous applications because they benefit from different aspects of GANs. 
In fact, the role of the discriminator is highlighted in detection methods. 
%benefit
The aforementioned methods show good performance in anomaly detection while they reduce the number of training data significantly.
%limitation
However, varying datasets and metrics employed for the experiments challenge a fair comparison between the methods.

\begin{table}[!ht]
	\tiny
	\centering\renewcommand\cellalign{lc}
	\caption{Detection GAN-based methods in medical image processing.}
	\label{table:detect}
	\begin{adjustbox}{width=\textwidth}{%
		\begin{tabu} to \textwidth {@{}X[0.8l] X[0.3l] X[0.35l] X[0.25l] X[0.4l] X[0.8l] X[0.1l] X[0.2l]@{}}
			\toprule
			method 	& Arch & loss &	modality & dataset & performance & PR & code \\ 
			\hline
			
			\cite{kohl2017adversarial} & \makecell{U-Net,\\ GAN} & \makecell{MSE, GAN}	& \makecell{MRI \\(prostate)}               & \makecell{ (NCT) \\Heidelberg\\Train:188}        &  \makecell{Specificity=0.98$\pm$0.14\\DSC=0.41$\pm$0.28\\Sensitivity=0.55$\pm$0.36} &No &No          
			\\ 
			\hline
			
			\cite{udrea2017generative} & \makecell{cGAN, \\U-net}& -	& \makecell{Natural\\Skin} &    \makecell{unknown\\Train:2417\\Test:583}     & \makecell{(Subjective)\\Correct Detect = 0.914} &Yes &No         
			\\
			\hline
			
			\cite{Tuysuzoglu2018DeepAC} & GAN & \makecell{Adv, \\Local, \\Contour}	& Ultra-Sound (prostate)               & \makecell{unknown\\Train:4570\\Test:229}        &  DSC = 0.92$\pm$0.3 &Yes &No 
			\\
			\hline
			
			\makecell{\cite{schlegl2017unsupervised}\\ \textbf{AnoGAN}} & DCGAN & Adv		&  \makecell{SD-OCT \\scans}               & \makecell{unknown\\Train:270\\Test:20}        & \makecell{Precision= 0.8834\\Recall= 0.7277\\Sensitivity=0.7279\\Specificity=0.8928\\AUC=0.89} &Yes &No           
			\\ 
			\hline
			
			\cite{Chen2018UnsupervisedDO} & \makecell{AnoGAN,\\ WGAN-GP}  & \makecell{WGAN-GP, \\Regular}		&      \makecell{MRI\\(brain)}           & \makecell{BRATS\\Train:35\\Test:42}        & \makecell{AUC = 0.92} &No &No            
			\\ 
			\hline
			
			\makecell{\cite{baumgartner2017visual}\\ \textbf{VA-GAN}} & \makecell{WGAN, \\U-Net}	&  Adv & \makecell{MRI \\(brain)}   & \makecell{ADNI\\Train:80\\Test:20}        & NCC = 0.27 &Yes &Yes           
			\\ 
			\bottomrule
		\end{tabu}%
		}
	\end{adjustbox}
\end{table}

%% file: sections/applications/denoising.tex
\subsection{De-noising}
\label{sub:denoising}

%Introduction
Imaging in diagnostic radiology typically involves a trade-off between image contrast and radiation hazard. 
A higher contrast might lead to better diagnosis, but exposes the patient to unwanted excessive radiation, whereas reduced radiation exposure leads to lower contrast and lower Signal-to-Noise Ratios. 
Deep Learning has been successfully used to denoise low-contrast images and to enhance their resolution.
However, these methods tend to produce blurry images. 
GANs, known to facilitate the generation of sharp, realistic looking images, provide the means to mitigate this problem. 
A variety of works have recognized this potential and proposed different ways to exploit GANs for denoising images with perceptually higher quality.

For example, \cite{wolterink2017generative} propose to learn tissue texture information from a small amount of paired data and address the blurring effect using GANs. 
Similarly, \cite{wang2018conditional} propose to use the cGAN to remove metal artifacts from CT images. 
\cite{yang2018low} utilize a combination of the W-GAN and a perceptual loss to improve training stability without losing perceptual quality. 
The sharpness of the denoised image is also the factor that \cite{Yi2018SharpnessAwareLC} (Sharpness Aware Generative Adversarial Network - SAGAN) worked on (Figure \ref{fig:denoise4}).

%\begin{figure}
%	\centering
%	\includegraphics[width=0.7\linewidth]{Diagrams/yi2017sharpness}
%	\caption{SAGAN architecture \citep{Yi2018SharpnessAwareLC}}
%	\label{fig:denoise4}
%\end{figure}

%Conclusion
Table \ref{table:denoise} summarizes major GAN-based denoising methods. 
%open direction
While visually, the results look compelling, it seems that an adequate, objective metric to evaluate the strength of methods in preserving important medical information of the image is not available yet. 
As PSNR, MSE, SSIM, SD, and mean - the most commonly used metrics in the evaluation of de-noising methods - are not sensitive enough to recognize texture details, the RoI of any image should be segmented to be measured by metrics, which is an expensive procedure. 
The introduction of a new, meaningful metric is subject to future work.
%advantages
Despite this limitation, visually results achieved in reviewed papers benefit from the ability of GANs in learning the main general features of a domain of images.
Also by modifying the loss function to consider more textural features, good performance in medical image de-noising is demonstrated. 
%limitation
However, finding a faster and more accurate framework is an open direction to be worked on in the future.

\begin{table}[]
	\tiny
	\centering\renewcommand\cellalign{lc}
	\caption{De-noising GAN-based methods in medical image processing}
	\label{table:denoise}
	\begin{adjustbox}{width=\textwidth}{%
		\begin{tabu} to \textwidth {@{}X[0.55l] X[0.15l] X[0.2l] X[0.2l] X[0.35l] X[0.8l] X[0.1l] X[0.2l]@{}}
			\toprule
			method 	& Arch & loss &	modality & dataset & performance & PR & code \\ \hline
			\cite{wolterink2017generative} & \makecell{CNN, \\GAN} & \makecell{CNN, \\Adv} 		&  \makecell{CT \\(phantom)\\ (cardiac)}               & \makecell{unknown\\Train:48\\Train:28}        & \makecell{Agatston Score:\\Median=20.7\\ Min=6.1\\Max=145.1}  &YEs &No          \\ \hline
			
			\cite{wang2018conditional} & \makecell{U-net, \\ResNet, \\cGAN} & \makecell{Adv, \\$L1$} 		&  \makecell{CT \\(ear)}               & \makecell{unknown\\Train:14346\\Test:74}       & \makecell{P2PEs:\\Median=0.409\\ STD=0.133\\Max=0.912}  &Yes &No          \\ \hline
			
			\cite{yang2018low} & \makecell{WGAN, \\VGG} & \makecell{features \\distance,\\WGAN}	&  CT              & \makecell{unknown\\Train:4000}        & \makecell{Noise Suppression=3.20$\pm$0.25\\Artifact Reduction=3.45$\pm$0.25\\Overall Quality=3.70$\pm$0.15}   &Yes &No         \\ \hline
			
			\makecell{\cite{Yi2018SharpnessAwareLC}\\ \textbf{SAGAN}} & \makecell{MGAN,\\ResNet} & \makecell{Pixel-wise, \\MGAN,\\Sharpness} 	& CT               & \makecell{CT phantom\\(Catphan 600)\\Train:4$\times$708\\Test:4$\times$142}        &  \makecell{(N=$10^4$)\\PSNR=26.77\\SSIM=0.8454\\(N=$10^5$)\\PSNR=28.25\\SSIM=0.87}  &Yes &No         \\ \bottomrule
		\end{tabu}%
		}
	\end{adjustbox}
\end{table}

%% file: sections/applications/Registration.tex
\subsection{Registration}
\label{sub:Registration}

%introduction
%Image registration is another application in medical image processing which contributes to align different medical images and extract combined diagnostic information. 
Traditional registration methods suffer from parameter dependency and heavy optimization load. CNNs have been successfully used to align medical images in a single forward-pass through the network.
Therein, GANs with their excellent image transformation capabilities have emerged as a candidate to extract a more optimal registration mapping.

\cite{fan2018adversarial} propose an unsupervised GAN to register structural patterns (defined in patches) among different brain images. 
In 3D prostate MRI and intra-procedural Transrectal Ultrasound (TRUS), \cite{Yan2018AdversarialIR} and \cite{Hu2018AdversarialDR} propose GANs for registration and deformation correction respectively. 
In the method proposed by \cite{Yan2018AdversarialIR}, the discriminator serves as a certainty evaluator during testing.

%conclusion
Table \ref{table:reg} summarizes properties of the mentioned methods and their performance. 
%benefit
In registration tasks both local and global features are of importance. Since GANs learn features at various scales to model the discrepancy between distributions, they provide this beneficial information.
%limitation
While GANs significantly enhance the performance of registration methods, in some real medical settings the required performance can still not be reached.

\begin{table}[!ht]
	\tiny
	\centering\renewcommand\cellalign{lc}
	\caption{Registration GAN-based methods in medical image processing}
	\label{table:reg}
	\begin{adjustbox}{width=\textwidth}{%
		\begin{tabu} to \textwidth {@{}X[0.5l] X[0.3l] X[0.3l] X[0.5l] X[0.3l] X[0.6l] X[0.1l] X[0.2l]@{}}
			\toprule
			method 	& Arch & loss &	modality & dataset & performance & PR & code \\ \hline
			
			\cite{fan2018adversarial} & \makecell{U-net, \\GAN} & \makecell{Adv, \\Regular}	&  \makecell{MRI 3D \\Brain}            & \makecell{LPBA40,\\ IBSR18,\\ CUMC12,\\ MGH10\\Train:30\\Test:10}        & \makecell{DSC=71.8$\pm$2.3$\%$\\DSC=57.8$\pm$2.7$\%$\\DSC=54.4$\pm$2.9$\%$\\DSC = 61.7$\pm$2.1$\%$\\\\\\}  &Yes &No       
			\\ \hline
			
			\cite{Yan2018AdversarialIR} & \makecell{CNN, \\WGAN} & Adv	&  \makecell{Prostate 3D\\ MRI and TRUS}              & \makecell{unknown \\Train:636\\Test:127}       & \makecell{TRE = 3.84 mm\\DSC = 0.58}   &Yes &No      
			\\ \hline
			
			\cite{Hu2018AdversarialDR} & 3D GAN & \makecell{Adv, \\DSC,\\Regular}	&  \makecell{Prostate 3D\\ MRI\\ TRUS} & \makecell{unknown\\108 pairs}	& \makecell{TRE = 6.3 mm\\DSC = 0.82} &Yes &No
			\\
			
			\bottomrule
		\end{tabu}%
		}
	\end{adjustbox}
\end{table}

%% file: sections/applications/classification.tex
\subsection{Classification}
\label{sub:classification}

%Introduction
CNNs, known as the best performing classifiers, require a huge amount of training data. However, difficulties in access to medical data hampers their effective employment. The ability of GANs to augment the training data and to extract domain-specific features of every class can potentially help to overcome this limitation.

For separating useful from non-informative images in cardiac Ultra-Sound (US), \cite{zhang2017semi} propose the Semi-Coupled GAN (SCGAN). 
In this work, two generators are employed to generate useful and non-informative cardiac samples; The discriminator is extended to differentiate both real/fake and useful/non-informative samples (Figure \ref{fig:class1}).

%\begin{figure}
%	\centering
%	\includegraphics[width=0.7\linewidth]{Diagrams/zhang2017semi}
%	\caption{SCGAN architecture \citep{zhang2017semi}}
%	\label{fig:class1}
%\end{figure} 

For classification of prostate histopathology images into two Gleason grades, \cite{Ren2018AdversarialDA} propose a GAN to extract features of every class and classify input patches based on those features. 
They also introduce a classification loss to prevent the network from classifying patches of the same image in the different Gleason grades.

%Conclusion
Table \ref{table:class} provides insights into the performance and summarizes properties of the mentioned methods. 
%benefit
The results indicate that the classification tasks benefit from the GAN-generated samples.
%limitation
	
\begin{table}[!ht]
	\tiny
	\centering\renewcommand\cellalign{lc}
	\caption{Classification GAN-based methods in medical image processing}
	\label{table:class}
	\begin{adjustbox}{width=\textwidth}{%
		\begin{tabu} to \textwidth {@{}X[0.5l] X[0.42l] X[0.25l] X[0.32l] X[0.12l] X[0.6l] X[0.1l] X[0.15l]@{}}
			\toprule
			method 	& Arch & loss &	modality & dataset & performance & PR & code \\ \hline
			
			\makecell{\cite{zhang2017semi}\\ \textbf{SCGAN}} & GAN & Adv	&  \makecell{MR \\(Cardiac)}              & \makecell{UKBB\\3400}        & \makecell{(MAS - MBS)\\Acc=92.5, 89.3\\Precision=87.6, 89.1\\Recall=90.5, 91.7} &Yes &No         
			\\ \hline
			
			\cite{Ren2018AdversarialDA} & \makecell{\cite{Su2013CellSI},\\GAN} & \makecell{Adv, \\patch-class}	&  \makecell{prostate \\histopathology} & \makecell{TCGA, \\CINJ}	& \makecell{(TCGA/TCGA to CINJ)\\Accuracy = 0.77, 0.75} &Yes &No
			\\
			
			\bottomrule
		\end{tabu}%
		}
	\end{adjustbox}
\end{table}

%% file: sections/discussion.tex
\section{Discussion}
\label{sec:discussion}

%Overview
GANs have been receiving significant attention from the medical imaging community - this is evident from the sudden spike in the number of papers published using GANs.

\subsection{Advantages}
\label{sub:advantages}
They are capable of mimicking data distributions, producing realistic looking images and learning rich similarity metrics - which are beneficial for discriminative deep learning frameworks.

\textbf{Scarcity of labeled data and class imbalance: }
Often times, annotations are expensive to obtain in medical imaging. 
This impairs supervised deep learning methods. 
At the same time, medical data often suffers from class imbalance, due to the rare nature of some pathologies. 
The ability of GAN in generating realistic looking images not only can be employed to enlarge training datasets but also it can be exploited in semi- and unsupervised settings. 
%Proposed methods using these advantages are reviewed in section \ref{sub:unsupervised_image_synthesis}, and \ref{sub:reconstruction}.
Great success has been made in image-to-image translation for cross-modality image synthesis, such as mapping from MR to CT data.
By synthesizing across modalities, conditional GANs reduce the cost and risk factors of medical image acquisition (Section \ref{sub:conditional_image_synthesis}), and can offer multi-modal information which can be fused for better diagnostic decision making.
In Unsupervised realms, the GAN's ability to learn data distributions opens up the possibility to detect unseen abnormal cases in real datasets (Section \ref{sub:detection}).

\textbf{Rich feature extraction:}
Learning distinctive patterns in medical images plays a vital role in the diagnosis of diseases. 
GANs, equipped with the adversarial training concept, are powerful in extracting semantically meaningful features from images which traditional pixel-wise losses fail to grasp. 
This property has been successfully leveraged in segmentation (Section \ref{sub:Segmentation}), registration (Section \ref{sub:Registration}), and de-noising (Section \ref{sub:denoising}).
On the other hand, the hierarchical discriminative potential of GANs has also been utilized for classification applications (Section \ref{sub:classification}).

% the best performances (where GAN solved the problem)

\subsection{Drawbacks}
\label{drawbacks}
We identify three major drawbacks in the current form of GANs that might hinder their acceptance in the medical community:

\textbf{Trustability of the Generated Data:} In healthcare, where gaining the clinicians trust is the biggest challenge for any technology, images generated by GANs provide little comfort.
The basic networks (generator and discriminator) are still deep neural networks, the mechanism of which is not sufficiently understood. 
In medical images intensities are typically associated with some meanings, e.g. in CT data every intensity can be mapped to the Houndsfield-scale and thus characterizes certain tissue.
Such an association and mapping is currently missing from the GAN reconstruction - a shortcoming severe enough for clinicians to distrust images synthesized with GANs.
In Computer Vision, where the overall perception is the main concern, these results are more adequate.

\textbf{Unstable Training:} The typical GAN training is unstable because of numerical and conceptual reasons \citep{creswell2018generative}.
This may result in convergence issues such as mode collapse or mode-hopping.
Fundamental theoretical work focuses on solving this problem and usually provides benchmarks on Computer Vision Datasets, where generated images are easy to interpret.
However, in medical imaging, where the modes of images are unclear, the identification of such unstable situations and unrealistic results can be very challenging.
In the reviewed methods, architectural modifications and customized loss functions are proposed to address these problems.
Yet, there is a lack of proper evaluation tools and medical benchmarks which measure the efficacy of the proposed methods in a comparable way.

\textbf{Evaluation:} In medical imaging, researchers mostly rely on traditional pixel-wise metrics to evaluate GANs performance.
This is tricky since GANs overcome the limitations of these traditional measures.
Most of these metrics are usable only when ground truth images are available and this challenges the evaluation of unsupervised methods.
On the other hand, as \cite{lucic2018gans} mentioned, specific metrics should be employed to evaluate the performance of GAN-based methods due to the randomness of the initialization/optimization procedure and model instability problem.
In the reviewed papers such a metric is not explored.

\textbf{Uninterpretability:} A model will be reliable in a medical environment if it follows features that clinicians consider in diagnosis and prognosis.
Even if a model provides information which is not simply recognized by medical experts, reasons for its decision should be interpretable.
While GANs show superior performance in many applications, they suffer from the same un-interpretability as other deep models.
This is the main obstacle to their practical application in medical environments.
Generally, Solving the interpretability of deep networks would open up a new direction to discover shortages in the models.

We believe GANs need to address the significant drawbacks discussed before being considered a trustworthy technology in practical medical image analysis. 
To this end, we can think of GANs as a technical building block rather than a stand-alone piece of technology for the future. 
For example, in the case of synthesizing CT data, enveloping GANs synthesis with a physics-based simulation might ensure realistic HU values.  
Training instabilities need to be addressed as well, which means rigorous experimentation to understand the convergence of GANs in the medical imaging context. 
Important steps have already been made in the Computer Vision field.
In short, along with exciting results, GANs open up many possible research questions for the next few years.
Proper understanding and answering those hold the key to their successful deployment in the real clinical scenario.